\documentclass[10pt,twocolumn,letterpaper]{article}

\usepackage{iccv}
\usepackage{times}
\usepackage{epsfig}
\usepackage{graphicx}
\usepackage{amsmath}
\usepackage{amssymb}
\usepackage{comment}

\usepackage{booktabs}
\usepackage{multirow}
\usepackage{array}
\usepackage[colorinlistoftodos]{todonotes}
\usepackage{bbm} %
\usepackage[normalem]{ulem} %

\usepackage{fancyhdr}

\graphicspath{{figures/}} %

\usepackage[pagebackref=true,breaklinks=true,letterpaper=true,colorlinks,bookmarks=false]{hyperref}

\let\oldenumerate\enumerate
\renewcommand{\enumerate}{
  \vspace{-6pt}
  \oldenumerate
  \setlength{\itemsep}{1pt}
  \setlength{\parskip}{0pt}
  \setlength{\parsep}{0pt}
}

\let\oldendenumerate\endenumerate
\renewcommand{\endenumerate}{
  \oldendenumerate
  \vspace*{-6pt}
}

\let\olditemize\itemize
\renewcommand{\itemize}{
  \vspace{-6pt}
  \olditemize
  \setlength{\itemsep}{3pt}
  \setlength{\parskip}{0pt}
  \setlength{\parsep}{0pt}
}

\let\oldenditemize\enditemize
\renewcommand{\enditemize}{
  \oldenditemize
 \vspace*{-6pt}
}

\definecolor{darkred}{rgb}{0.6,0,0}
\definecolor{darkgreen}{rgb}{0,0.3,0}
\definecolor{darkblue}{rgb}{0,0,0.5}

\iccvfinalcopy %

\def\iccvPaperID{4024} %

\begin{document}

\title{Phrase Localization Without Paired Training Examples}

\author{Josiah Wang\\
Imperial College London\\
{\tt\small http://www.josiahwang.com}
\and 
Lucia Specia\\
Imperial College London\\
{\tt\small l.specia@imperial.ac.uk}
\thanks{* The authors thank the anonymous reviewers and area chairs for their feedback. This work was supported by the MultiMT project (H2020 ERC Starting Grant No. 678017); and by the MMVC project, via an Institutional Links grant (ID 352343575) under the Newton-Katip \c{C}elebi Fund partnership funded by the UK Department of Business, Energy and Industrial Strategy (BEIS) and Scientific and Technological Research Council of Turkey (T{\"U}B{\.I}TAK) and delivered by the British Council. }
}

\renewcommand\footnotemark{}

\maketitle
\thispagestyle{fancy}

\begin{abstract}
Localizing phrases in images is an important part of image understanding and can be useful in many applications that require mappings between textual and visual information. Existing work attempts to learn these mappings from examples of phrase-image region correspondences (strong supervision) or from phrase-image pairs (weak supervision). We postulate that such paired annotations are unnecessary, and propose the first method for the phrase localization problem where neither training procedure nor paired, task-specific data is required. Our method is simple but effective: we use off-the-shelf approaches to detect objects, scenes and colours in images, and explore different approaches to measure semantic similarity between the categories of detected visual elements and words in phrases. Experiments on two well-known phrase localization datasets show that this approach surpasses all weakly supervised methods by a large margin and performs very competitively to strongly supervised methods, and can thus be considered a strong baseline to the task. The non-paired nature of our method makes it applicable to any domain and where no paired phrase localization annotation is available.
\end{abstract}

\section{Introduction}
\label{sec:introduction}

\begin{figure}[!t]
\centering
\includegraphics[width=0.90\linewidth]{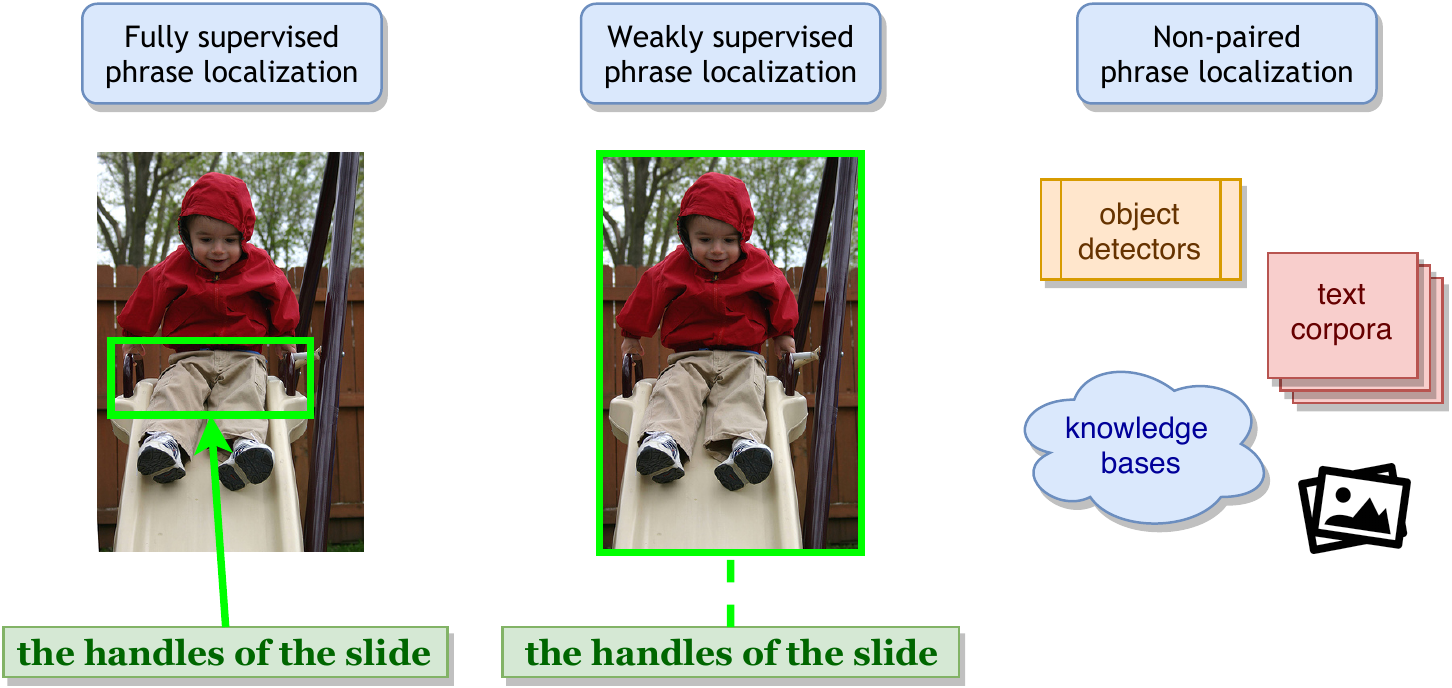}
   \caption{We investigate the task of phrase localization \emph{without} paired training examples. Conventional settings require phrase and image localization annotations (fully supervised) or phrase and image pairs (weakly supervised) at training time. In contrast, the non-paired setting does not provide such annotations for training, but instead allows models to exploit resources such as off-the-shelf visual detectors, large-scale general corpora, knowledge bases and generic images to localize previously unseen phrases at test time. This non-paired setting is thus a baseline to supervised settings.
   }
\label{fig:overview}
\end{figure}

Significant progress has been made in recent years in the task of detecting and localizing instances of object categories in images, especially with deep convolutional neural network (CNN) approaches to object detection~\cite{Girshick:2015,GirshickEtAl:2014,HeEtAl:2015,LiuEtAl:2016,RedmonEtAl:2016,RedmonFarhadi:2017,RenEtAl:2015,SermanetEtAl:2014}. In most work, object detection labels are treated as a fixed set of category labels, and visual detectors are trained to localize each category in the image. 
In more realistic applications, however, people %
refer to objects in images via free-form textual phrases, instead of object categories. For example, \emph{a brown and furry puppy} instead of \emph{dog}. Phrase-level localization has been introduced~\cite{HinamiSatoh:2018,KazemzadehEtAl:2014,KrishnaEtAl:2017,MaoEtAl:2016,PlummerEtAl:2015,YuEtAl:2016} to address this need by combining visual object recognition and natural language processing. What we refer to as `phrases' can include single words, short clauses or phrases, or even complete sentences. All previous work assume some form of supervision at training time: either strong supervision (object localization for the phrase in the image is provided) \cite{ChenEtAl:2018b,ChenEtAl:2017,HinamiSatoh:2018,HuEtAl:2016,PlummerEtAl:2018,PlummerEtAl:2017b,RohrbachEtAl:2016,WangEtAl:2016a,ZhangEtAl:2017} or weak supervision (the phrase and image pair is provided, but not the object's localization in the image) \cite{ChenEtAl:2018,XiaoEtAl:2017,YehEtAl:2018,ZhaoEtAl:2018} (Figure~\ref{fig:overview}). Such specific bounding box annotations and even image-phrase pairs, however, are hard and labourious to obtain. This makes it difficult to scale detection up to more realistic settings covering the large space of possible phrases that can be uttered by a person. 

In this paper, we tackle
the novel task of \textbf{phrase localization in images \emph{without} any paired examples}, \ie the model has access to neither phrase-image pairs nor their localization in the image at training time (a `training' phase may not even be required).
To our knowledge, no previous work has explored this challenging setting of performing phrase localization without paired annotations (image-level or object-level). We argue that such a `non-paired' setting better reflects how humans localize objects in images -- not by memorizing paired examples, but by assembling prior knowledge from %
more general sources and tasks (\eg recognizing concepts or attributes) to tackle a more specialized task (phrase localization).  Thus, this setting acts as a strong baseline to the phrase localization task, \ie it demonstrates the extent to which a system can perform phrase localization even without having seen any such examples. This can give further insights into how paired examples can be better utilized for phrase localization in an informed manner, on top of what can be done without paired examples. The approach is also scalable to any domain and to any number of natural language and image pairs.

The main contribution of this paper is %
a model for phrase localization that is \emph{not} trained on phrase localization annotations (Section~\ref{sec:framework}).
Instead, it exploits readily available resources, tools and external knowledge. Our model has the advantage of being \textbf{simple} and \textbf{interpretable}, acting as a strong baseline for the novel, non-paired setting. 
We provide an in-depth analysis of this model on two existing phrase localization datasets (Section~\ref{sec:results}), using different detectors and combination of detectors, semantic similarity measures for concept selection, and strategies to combine these components to localize previously unseen phrases. 

Our experiments on two existing phrase localization datasets show that our approach without paired examples outperforms state-of-the-art weakly supervised models by a large margin, and is on par with fully supervised approaches that utilize large sets of annotated phrase localization examples and domain-specific tools at training time. The results suggest that, for these datasets, training with phrase localization annotations may not be necessary or optimal for tackling the phrase localization task.

\section{Related work}
\label{sec:relatedwork}

\begin{figure*}[!t]
  \centering
    \includegraphics[width=0.70\linewidth]{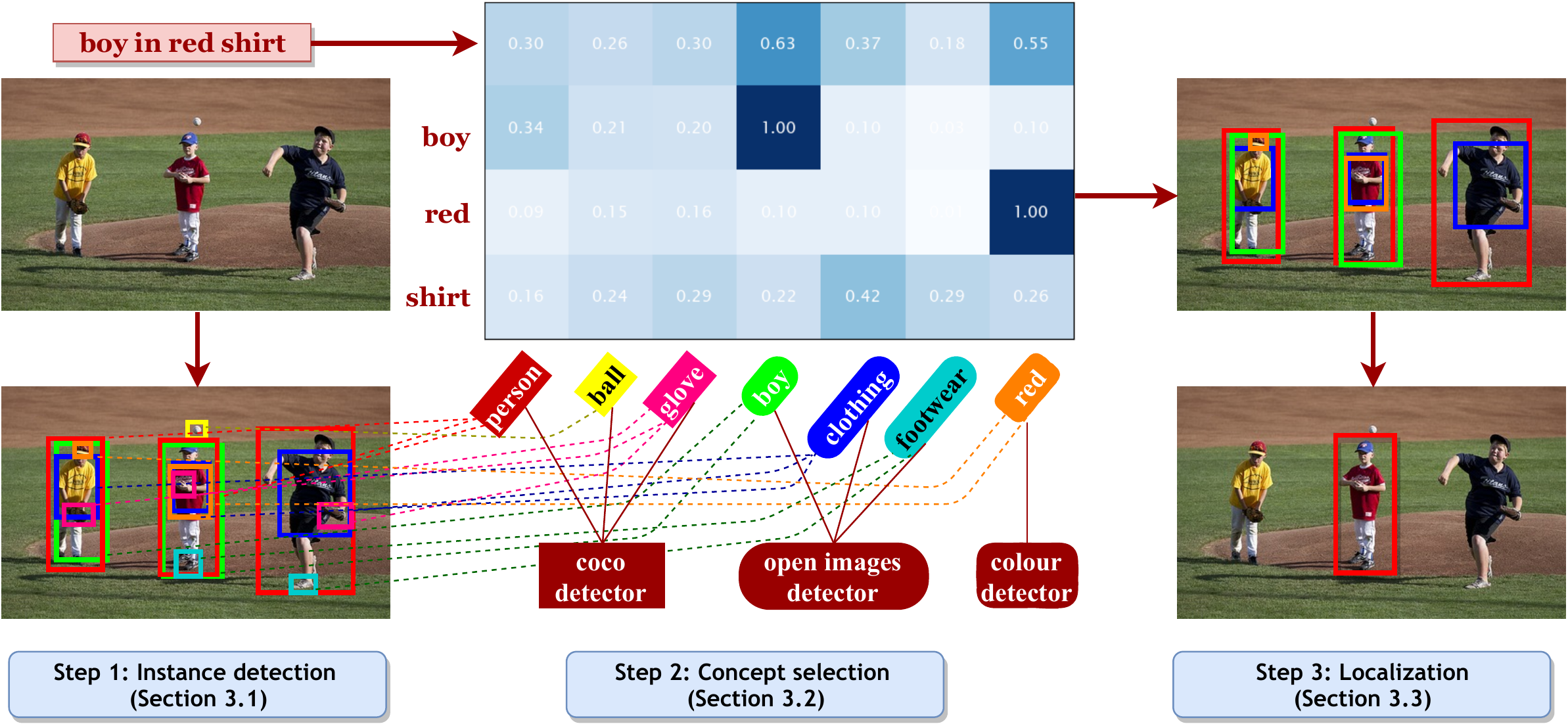}
   \caption{The three stages of the proposed model for non-paired phrase localization. The \textbf{instance detection} phase detects instances of various concepts using pre-trained detectors. The \textbf{concept selection} stage ranks these detected concepts against the query phrase (using pre-trained word embeddings) and forwards the best candidate concept instance(s) to the \textbf{localization} phase, where the model predicts the final bounding box for the query phrase.}
\label{fig:model}
\end{figure*}

The availability of datasets annotated with bounding box labels~\cite{ChenEtAl:2015,RussakovskyEtAl:2015} has allowed the development of deep CNN based detectors~\cite{Girshick:2015,GirshickEtAl:2014,HeEtAl:2015,LiuEtAl:2016,RedmonEtAl:2016,RedmonFarhadi:2017,RenEtAl:2015,SermanetEtAl:2014}, propelling the field of object recognition to produce more accurate detections and localization of object instances in images. %

There has been recent interest in localizing objects using free-form natural language phrases instead of fixed labels, with datasets constructed for such tasks~\cite{KazemzadehEtAl:2014,KrishnaEtAl:2017,MaoEtAl:2016,PlummerEtAl:2015,YuEtAl:2016}. We classify existing phrase localization or grounding approaches as either strongly/fully supervised~\cite{ChenEtAl:2018b,ChenEtAl:2017,HinamiSatoh:2018,HuEtAl:2016,PlummerEtAl:2018,PlummerEtAl:2017b,RohrbachEtAl:2016,WangEtAl:2016a,ZhangEtAl:2017} or weakly supervised~\cite{ChenEtAl:2018,KarpathyEtAl:2014,XiaoEtAl:2017,YehEtAl:2018,ZhaoEtAl:2018}. %

Methods that use strong supervision include those that project phrases and image regions onto a common space~\cite{PlummerEtAl:2017b,PlummerEtAl:2017,WangEtAl:2016a}, those that build a language model for the phrase conditioned on bounding box proposals~\cite{HuEtAl:2016}, and those that learn to attend to the correct region proposals given the phrase~\cite{RohrbachEtAl:2016}. More recent approaches include conditioning an object detector %
on phrases instead of fixed object labels~\cite{HinamiSatoh:2018}, leveraging semantic context and learning to regress bounding boxes directly from phrase localization data instead of relying on external region proposals~\cite{ChenEtAl:2018b,ChenEtAl:2017}, and conditioning %
embeddings on the categories/groups to which a phrase belongs~\cite{PlummerEtAl:2018}.

In a weakly supervised setting, no localization is provided at training time. Thus, such methods use external region proposals~\cite{ChenEtAl:2018,ZhaoEtAl:2018}, generic object category detectors~\cite{YehEtAl:2018} and attention maps~\cite{XiaoEtAl:2017} for localization. To learn to associate these region proposals with phrases, various methods have been proposed, including learning to constrain the regions' spatial positions using the parse tree of the caption~\cite{XiaoEtAl:2017}, performing a continuous search using region proposals as anchors~\cite{ZhaoEtAl:2018}, linking words in text and detection labels using co-occurrence statistics from paired captions~\cite{YehEtAl:2018}, and enforcing consistency between the concept labels of a region proposal and words in the query~\cite{ChenEtAl:2018}.

We are unaware of work that tackles phrase localization without paired examples. Yeh~\etal~\cite{YehEtAl:2018} define their work as `unsupervised', but we consider it weakly supervised. Their model uses image-phrase pairs from the training dataset (similar distribution as the test set) to compute co-occurrence statistics between words and concepts, and to train %
image classifiers for words in the phrases. %
Our model adapts Yeh~\etal's approach for when no paired training examples are available.
In addition, we propose a new localization module that makes better direct use of the output of multiple detectors for phrase localization.

\section{Model for non-paired phrase localization}
\label{sec:framework}

\paragraph{Task definition. } Given an image $I$ and a query phrase $q$ at \emph{test} time, the aim of the phrase localization task is to produce a bounding box $b$ encompassing the visual entity in $I$ to which $q$ refers. %
In contrast to conventional supervised settings, in our proposed non-paired setting annotated paired training examples ($q$,$I$) or ($q$,$I$,$b$) are \emph{not} available at training or model construction time. Instead, models are allowed to use external resources that are not specific to phrase localization, for example general visual object detectors, generic text corpora, knowledge bases and thesauri, and images from generic datasets not annotated with phrases. We note that visual detectors may be trained in a supervised manner (\eg with COCO or ImageNet), but there is no supervision in terms of phrase-based labels for the phrase localization task. %
Similarly, language models trained from generic text corpora may contain phrases from the test set, as long as they are independent of the images.

Our model builds upon the approach of Yeh~\etal~\cite{YehEtAl:2018}. In contrast to their approach, however, we perform phrase localization without an explicit training step or phrase localization annotations. We (i) incorporate a semantic similarity measure derived from general text corpora rather than aligned training examples; (ii) explore an array of off-the-shelf visual detectors %
not specifically trained for phrase localization; (iii) propose different strategies to perform phrase localization from detection outputs, including a novel consensus-based method that combines the output of multiple detectors.

At test time, our model performs phrase localization using a three-step process (Figure~\ref{fig:model}). 
In the first step -- \textbf{instance detection} -- it predicts candidate bounding boxes %
using a combination of different visual detectors (Section~\ref{sec:instancedetection}). In the second step -- \textbf{concept selection} -- the model computes the semantic similarity between the query phrase and the concept labels for instances detected in the previous step, and selects the most relevant  instance(s) %
(Section~\ref{sec:conceptselection}). In the final step -- \textbf{localization} -- the model predicts the bounding box for the query phrase from the selected candidate instance(s) from the second step %
(Section~\ref{sec:localization}).

\subsection{Instance detection}
\label{sec:instancedetection}

The first stage of our non-paired phrase localization model relies on different visual object detectors. We explore using the detectors in isolation and by combining their output, where the concepts are not necessarily mutually exclusive. The key idea is to exploit the redundancy from multiple detectors to handle missing detections and to increase the importance of object instances detected across multiple detector groups. We experiment with the following:

\begin{enumerate}
\item \textbf{tfcoco}: A Faster R-CNN~\cite{RenEtAl:2015} detector trained %
to detect the $80$ categories of MS COCO~\cite{LinEtAl:2014}, using the Tensorflow Object Detection API~\cite{HuangEtAl:2017},\footnote{\texttt{faster\textunderscore rcnn\textunderscore inception\textunderscore resnet\textunderscore v2\textunderscore atrous\textunderscore coco}} %
with confidence threshold of $0.1$.
\item \textbf{tfcoco20}: A subset of \textbf{tfcoco}, where we only consider the subset of $20$ categories from PASCAL VOC~\cite{EveringhamEtAl:2010}. This enables comparison to previous work. %
\item \textbf{tfoid}: Another Faster R-CNN detector, trained to detect the $545$ object categories of the Open Images Dataset (V2)~\cite{KrasinEtAl:2017}, again using the TensorFlow Object Detection API,\footnote{\texttt{faster\textunderscore rcnn\textunderscore inception\textunderscore resnet\textunderscore v2\textunderscore atrous\textunderscore oid}} with confidence threshold of $0.1$.
\item \textbf{places365}: A WideResNet18~\cite{ZagoruykoKomodakis:2016} classifier trained on the Places2 dataset \cite{ZhouEtAl:2018} for $365$ scene categories. We assume that scenes usually cover the full image, and return the whole image as the bounding box localization when the classification confidence is at least $0.1$. We keep only the top $20$ predicted classes.
\item \textbf{yolo9000}: A YOLO9000 detector \cite{RedmonFarhadi:2017} trained on MS COCO and ILSVRC~\cite{RussakovskyEtAl:2015} for $9,413$ categories in a weakly supervised fashion. We use YOLOv2.
\item \textbf{colour}: A colour detector for $11$ basic English colour terms, derived from the posterior across the colour terms for RGB pixels as learned from real world images~\cite{vandeWeijerEtAl:2007}. We performed connected component labelling (8-connectivity) after thresholding the posteriors at $0.3$ and generated bounding boxes for each labelled connected component. The area of the bounding boxes is constrained to be at least $625$ pixels.
\end{enumerate}

The detectors vary in accuracy and the number and type of categories covered. It is worth noting that none of the detectors above directly use images or phrase localization annotations from our test datasets. %
This will emphasize the ability of our phrase localization model to generalize to unseen data. 
More detectors could potentially be used to further improve recall, but the ones used here are sufficient to show that the proposed approach is very promising.

\subsection{Concept selection}
\label{sec:conceptselection}

The second stage of our model bridges the query phrase to be localized and the output of detectors from Section~\ref{sec:instancedetection}. It computes the semantic similarity between each phrase and the detector concept labels. The intuition is that the detected instance of a concept that is very similar or related to a word or phrase in the query is most likely to be the target object. For example, the word \emph{dancer} might be highly similar or related to the category \emph{person}; thus even without a \emph{dancer} detector, the model can infer that the detected \emph{person} is likely to be the \emph{dancer} mentioned in the query. 

We represent both queries $q$ and concept labels $c$ as $300$-dimensional CBOW word2vec embeddings~\cite{MikolovEtAl:2013}. %
Multiword phrases are represented by the sum of the word vectors of each in-vocabulary %
word of the phrase, normalized to unit vector  by its $L_2$-\emph{norm}.\footnote{Averaging word embeddings for the entire phrase leads to the same experimental results.} 
All words in the queries and concept labels (except \textbf{yolo9000}) are lowercased. %
For \textbf{yolo9000}, each category is a WordNet~\cite{Fellbaum:1998} synset. Thus we represent each category as the sum of the word vectors for each term in its synset, normalized to unit vector. Out-of-vocabulary words are addressed by matching case variants of the words (\emph{Scotch whiskey} to \emph{scotch whiskey}). %
Failing that, we attempt to match multiword phrases like before. %

We noticed many misspellings among the query phrases. %
Thus, the model exploits another external resource to perform automated spell correction\footnote{\url{https://pypi.org/project/pyspellchecker/}} for out-of-vocabulary words. The model finds candidate replacement words from word2vec's vocabulary and choosing the one with the highest frequency in the corpus used to train the %
embeddings. The model consistently obtained slightly higher accuracies with spell correction, and thus we report only the results with the spell-corrected queries. 

We explore two approaches to aggregate the words in query phrases: as a single vector by summing the word vectors and normalizing to unit vector (\textbf{w2v-avg}), or by representing each word individually (\textbf{w2v}) and using only one of the words for localization (see Section~\ref{sec:localization}).

We use cosine similarity as the semantic similarity measure $S(q,c)$ between a query $q$ and a concept label $c$. 
This stage outputs a ranked list of candidate bounding box detections based on their similarity to the query phrase.

\subsection{Localization}
\label{sec:localization}

In the final stage, our proposed model predicts a bounding box given the query phrase and the ranked list of candidate detections from Section~\ref{sec:conceptselection}. This is accomplished %
by selecting from or aggregating the candidate instances that are most semantically similar to the query.

The simplest localization approach is to select from candidate detections the %
concept most similar to the query phrase. %
Where multiple instances of the same concepts are detected%
, we experiment with different tie-breaking strategies: (i) selecting a \textbf{random} instance; (ii) selecting the instance with the \textbf{largest} bounding box; (iii) selecting the instance with the highest class prediction \textbf{confidence}; (iv) generating a minimal bounding box enclosing \emph{all} instances (\textbf{union}). The latter may useful for dealing with queries referring to multiple instances of an object (\eg localizing \emph{three people} from three individual \emph{person} detections). 

Besides simple heuristics, we also propose a novel tie-breaking approach by \textbf{consensus}. The main idea is that %
detectors can vote on the most likely localization, exploiting the redundancy across detectors and the different aspects of the phrase (\emph{blue shirt}). %
We consider the semantic similarity of instances from the top-$K$ concepts above a similarity threshold (we use $K$=$5$ and %
threshold $0.6$%
). For each concept $c_i$, a pixel-level heatmap for the image, $M_{c_i}(I)$ is generated by setting to $1$ pixels that overlap with any bounding box instance of the concept, and setting to $0$ those that do not. We generate a combined heatmap $\hat{M}(I)$ by summing the %
heatmaps for each concept, each weighted by the semantic similarity score $S(q,c)$ from Section~\ref{sec:conceptselection}:
\begin{equation}
\hat{M}(I) = \sum_{i=1}^{K} S(q,c_i) M_{c_i}(I) 
\label{eq:heatmap}
\end{equation}

Phrase localization is performed by selecting the bounding box instances that voted for the pixels with the highest values, and choosing the box with the highest semantic similarity score as the predicted localization. In cases where there are multiple top scoring boxes, the model predict a minimal bounding box that encloses all such boxes. 

We compare using a single combined word embedding for the phrase (\textbf{w2v-avg}) or using the embedding for one word to represent the phrase (\textbf{w2v}). For the latter, we can select the word with the highest semantic similarity to any detected concepts (\textbf{w2v-max}). Intuitively, we only consider one word from the phrase for localization, where this word has the highest similarity to a detected concept. Alternatively, we can use the \emph{last} word %
for localization (\textbf{w2v-last}), assuming the last word is the head word. We default to localizing to the whole image when no words in the phrase are found in the vocabulary.

\section{Experimental results}
\label{sec:results}

We evaluate our proposed %
models on two challenging datasets: \textbf{Flickr30kEntities} (Section~\ref{sec:results-flickr30k}) and \textbf{ReferItGame} (Section~\ref{sec:results-refclef}). Both have been used to evaluate supervised phrase localization~\cite{ChenEtAl:2018,RohrbachEtAl:2016,YehEtAl:2018}. Each dataset represent different challenges: Flickr30kEntities are noun phrases extracted from full image captions, while ReferItGame are short phrases generated from an interactive game where one player tries to localize the object the other player is describing. Thus, we consider the latter as more challenging. We also test selected models on Visual Genome (Section~\ref{sec:results-visgenome}) to investigate the model's scalability to a different dataset with sentence-level descriptions.

\subsection{Evaluation metric}
\label{sec:metric}

As in previous work, we use the \emph{accuracy} metric for evaluation\footnote{Our evaluation script can be found at \url{https://github.com/josiahwang/phraseloceval}.}, where a predicted bounding box $p_i$ for a query phrase is considered correct if its intersection over union (IoU) with the ground truth $g_i$ is at least $50\%$. 

For reference, we measure the extent to which the correct localization can be found among the candidate localizations from the concept selection stage (Section~\ref{sec:conceptselection}), depending on the similarity measure and the detector used. This \emph{upperbound} accuracy is computed across $N$ test instances as 

\begin{equation}
\frac{1}{N} \sum_{i=1}^{N} \max_{j=1}^{B} \mathbbm{1}{\big(IoU(g_i, b_j) \geq 0.5\big)}
\end{equation} 

\noindent where $\mathbbm{1}{(\cdot)}$ is the indicator function and $B$ the number of candidate bounding boxes. We report a version of the upperbound that additionally includes %
the minimal bounding box that encompasses the \emph{union} of all candidates (thus $B+1$ candidates). This variant consistently gave higher upperbound accuracies than without the union. The results of both variants are given in the supplementary document.

\subsection{Phrase localization on Flickr30kEntities}
\label{sec:results-flickr30k}

\textbf{Flickr30kEntities}~\cite{PlummerEtAl:2015} is based on Flickr30k~\cite{YoungEtAl:2014}, containing bounding box annotations for noun phrases occurring in the corresponding image captions. The test split~\cite{PlummerEtAl:2017} comprises $14,481$ phrases for $1,000$ images, which we use for evaluation. The training and validation splits are not used in our non-paired localization experiments.

As no non-paired phrase localization work exists, we compare our method against a baseline of always localizing to the whole image ($21.99\%$ accuracy), and compare our models using different detectors and localization strategies.

As reference, we also compare our model against 
supervised approaches trained in a fully~\cite{ChenEtAl:2017,HinamiSatoh:2018,PlummerEtAl:2017b,RohrbachEtAl:2016} or weakly~\cite{ChenEtAl:2018,YehEtAl:2018} supervised setting. %
Note that these systems are not directly comparable to ours. In fact, the comparison is unfavourable to us as these work also use external tools like visual detectors or bounding box proposal generators in addition to supervised phrase localization training data.

\begin{table}[t]
  \centering
  	\small
  	\begin{tabular}{l l c l}
    \toprule
    \textbf{Detector} & \textbf{Similarity} & \textbf{Strategy} & \textbf{Acc (UB) \%} \\
	\midrule
    \multicolumn{3}{l}{Baseline: Always localize to whole image} & 21.99\\
    CC+OI & - & largest & 30.32 (73.00)\\
    20 & w2v-avg & union & 36.49 (51.81)\\
    CC & w2v-max & union & 37.57 (51.22)\\
    OI & w2v-max & union & 44.69 (50.04)\\    
    CC+OI & w2v-max & union & 48.20 (55.85)\\
    CC+OI+PL+CL & w2v-avg & consensus & 49.51 (58.93)\\    
	CC+OI+PL+CL & w2v-avg & union & 49.61 (58.10)\\
    CC+OI+PL & w2v-avg & consensus & 50.11 (58.00)\\    
	CC+OI+PL+CL & w2v-last & union & 50.36 (57.81)\\    
    CC+OI+PL & w2v-max & union & \textbf{\underline{50.49}} (57.81)\\ 
    \midrule
    \multicolumn{4}{l}{\textbf{\emph{Weakly supervised}}}\\
    GroundeR~\cite{RohrbachEtAl:2016} & & & 28.94 \\    
    Yeh~\etal~\cite{YehEtAl:2018} & & & 36.93\\
    \multicolumn{2}{l}{KAC Net + Soft KBP~\cite{ChenEtAl:2018}} & & 38.71 \\
    \noalign{\medskip}
    \multicolumn{4}{l}{\textbf{\emph{Strongly supervised}}}\\
    GroundeR~\cite{RohrbachEtAl:2016} & & & 47.81 \\
    SPC+PPC~\cite{PlummerEtAl:2017b} & & & 55.85 \\   
    QRC Net~\cite{ChenEtAl:2017} & & & 65.14\\  
    \multicolumn{2}{l}{Query Adaptive R-CNN~\cite{HinamiSatoh:2018}} & & 65.21\\    
    \bottomrule
    \end{tabular}
\caption{Accuracies (and upperbound \textbf{UB}) %
of some of our selected models on Flickr30kEntities, comparing different detector combinations, semantic similarity measures and localization strategies. As a comparison against supervised settings, we present our results alongside selected strongly and weakly supervised systems. 
These systems are not directly comparable to ours as they use phrase localization annotations for training. %
Keys: CC=\textbf{tfcoco}, OI=\textbf{tfoid}, 20=\textbf{tfcoco20}, PL=\textbf{places365}, CL=\textbf{colour}.}
\label{tbl:results-flickr30k}
\end{table}

Table~\ref{tbl:results-flickr30k} shows the accuracies on Flickr30kEntities for bounding box predictions from a selection of our models, using different combination of detectors, concept selection and localization strategies. Our best performing model combines \textbf{tfcoco}, \textbf{tfoid} and \textbf{places365} detectors with the \textbf{w2v-max} concept selector and the \textbf{union} localization strategy. %
This model comfortably outperformed the state-of-the-art weakly supervised model~\cite{ChenEtAl:2018} on this dataset ($50.49\%$ vs.\ $38.71\%$). Its accuracy is also higher than a strongly supervised model~\cite{RohrbachEtAl:2016} ($47.81\%$) and is competitive against others~\cite{ChenEtAl:2017,HinamiSatoh:2018,PlummerEtAl:2017b} that use strong supervision along with specialised detectors for the dataset, %
part-of-speech taggers and parsers, the full caption, %
and %
takes into account other entities/relations mentioned in the caption. In contrast, our method is much simpler and does not rely on domain-specific paired training data. The full results with different detector combinations, concept selection and localization strategies are provided as supplementary material. These results suggest that paired %
annotations might not even be completely necessary for the task, at least for Flickr30kEntities.

\begin{table*}[t]
\centering
\resizebox{\linewidth}{!}{
\small
\begin{tabular}{lcccccccc|c}
\toprule
 & \textbf{people} & \textbf{clothing} & \textbf{bodyparts} & \textbf{animals} & \textbf{vehicles} & \textbf{instruments} & \textbf{scene} 
& \textbf{other} & \textbf{overall} \\
\midrule
\# instances & 5626 & 2306 & 523 & 518 & 400 & 162 & 1619 & 3374 & 14481 \\
\midrule
20 (max, u) & 60.31 & 9.63 & 2.10 & 82.43 & 74.75 & 19.14 & 17.85 & 17.96 & 36.33 \\
CC (max, u) & 56.35 & 10.45 & 1.72 & 83.59 & 79.25 & 17.90 & 15.69 & 29.79 & 37.57 \\
CC+OI (max, u) & \textit{66.34} & \textit{37.99} & 21.03 & \textbf{84.75} & 79.75 & \textit{47.53} & 20.14 & 33.11 & 48.20 \\
CC+OI+PL (avg, u) & 66.18 & 35.52 & 21.03 & \textbf{84.75} & \textbf{81.00} & \textit{47.53} & 39.16 & \textit{34.71} & 50.27 \\
CC+OI+PL (max, u) & 66.27 & 37.55 & 20.65 & \textbf{84.75} & 80.00 & \textit{47.53} & 38.91 & 34.41 & 50.49 \\
CC+OI+PL+CL (avg, u) & 65.22 & 35.65 & \textit{21.22} & 78.19 & 78.00 & \textit{47.53} & 40.58 & 34.05 & 49.61 \\
\midrule
\multicolumn{10}{l}{\textbf{\emph{Weakly supervised}}}\\
Yeh~\etal~\cite{YehEtAl:2018} & 58.37 & 14.87 & 2.29 & 68.91 & 55.00 & 22.22 & 24.87 & 20.77 & 20.91\\
KAC Net (Soft KBP)~\cite{ChenEtAl:2018} & 58.42 & 7.63 & 2.97 & 77.80 & 69.00  & 20.37 & 43.53 & 17.05 & 38.71 \\
\noalign{\medskip}
\multicolumn{10}{l}{\textbf{\emph{Strongly supervised}}}\\
SPC+PPC~\cite{PlummerEtAl:2017b} & 71.69 & 50.95 & 25.24 & 76.25 & 66.50 & 35.80 & 51.51 & 35.98 & 55.85 \\
QRC Net~\cite{ChenEtAl:2017} & 76.32 & 59.58 & 25.24 & 80.50 & 78.25 & 50.62 & 67.12 & 43.60 & 65.14 \\
Query Adaptive R-CNN~\cite{HinamiSatoh:2018} & 78.17 & 61.99 & 35.25 & 74.41 & 76.16 & 56.69 & 68.07 & 47.42 & 65.21\\
\bottomrule
\end{tabular}
}
\caption{Non-paired phrase localization accuracies for different phrase types, as defined in Flickr30kEntities. \textbf{Bolded} results show higher accuracies than strongly supervised models, while \textit{italicized} accuracies indicate that they are higher than weakly supervised models. Keys: CC=\textbf{tfcoco}, OI=\textbf{tfoid}, 20=\textbf{tfcoco20}, PL=\textbf{places365}, CL=\textbf{colour}, max=\textbf{w2v-max}, avg=\textbf{w2v-avg}, u=\textbf{union}.}
\label{tbl:catresults-flickr30k}
\end{table*}

Table~\ref{tbl:catresults-flickr30k} gives the per-category breakdown of the accuracies. Our best models resulted in higher accuracies than all strongly supervised models for two out of eight categories (animals and vehicles). Our models also achieved better accuracies than weakly supervised models in seven out of eight categories, and are competitive for the remaining category (scene) against KAC Net~\cite{ChenEtAl:2018} ($40.58\%$ vs.\ $43.53\%$) and outperformed Yeh~\etal~\cite{YehEtAl:2018} ($24.87\%$). %

\subsubsection{Discussion}

\begin{figure*}[!t]
  \centering
    \begin{tabular}{cccc}
skyscrapers & a blue swimsuit & three men & A very excited drummer\\
\includegraphics[height=2.3cm]{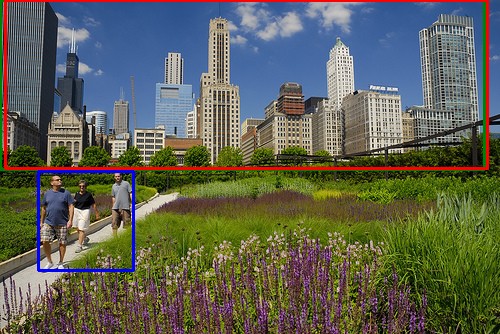} & \includegraphics[height=2.3cm]{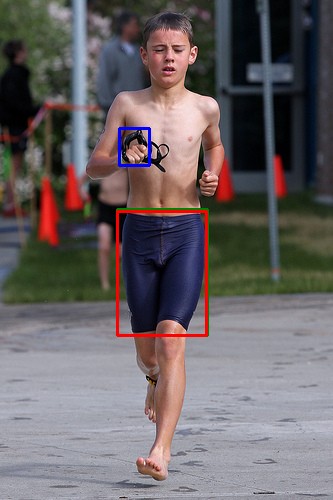} & \includegraphics[height=2.3cm]{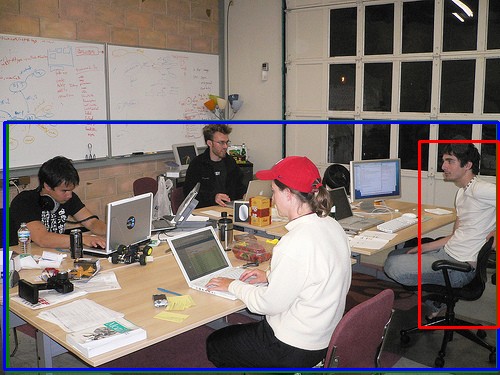} & \includegraphics[height=2.3cm]{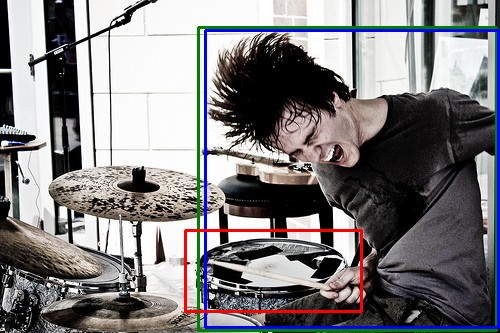}\\
    \noalign{\smallskip}
sky & lamp & soda & glass\\
\includegraphics[height=2.3cm]{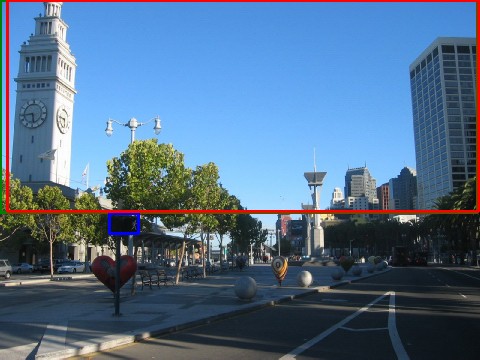} & \includegraphics[height=2.3cm]{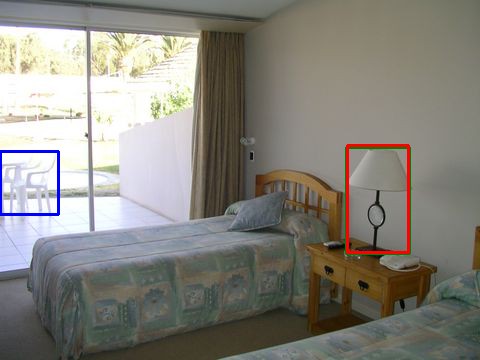} & \includegraphics[height=2.3cm]{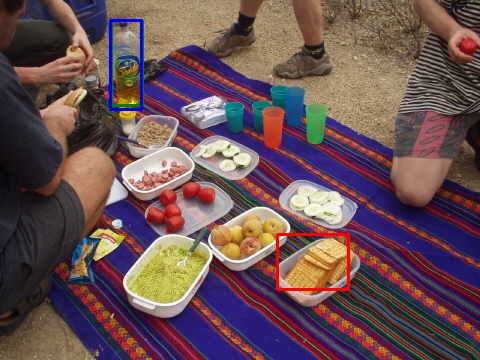} & \includegraphics[height=2.3cm]{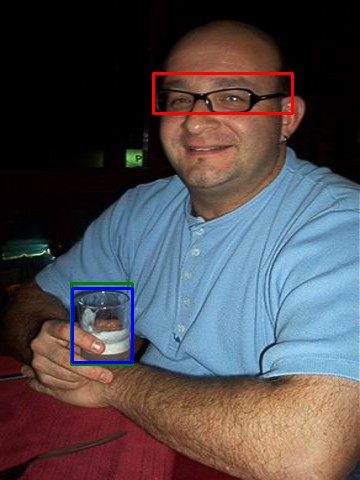}\\  
    \end{tabular}
   \caption{Example localization output for Flickr30kEntities (top row) and ReferItGame (bottom row). We compare the effects of adding a \textbf{tfoid} detector (\textcolor{darkred}{red} bounding box) to \textbf{tfcoco} (\textcolor{darkblue}{blue}) (\textbf{w2v-max}, \textbf{union}). The ground truth is indicated in \textcolor{darkgreen}{green}. The first two columns show examples of where adding a \textbf{tfoid} detector improves localization, while the last two columns are examples where it hurts localization.}
\label{fig:output-oid}
\end{figure*}

\begin{figure*}[!t]
  \centering
  \begin{tabular}{cccc}
a yellow tennis suit & a long green shirt & a red toy & A blue , red , and yellow airplane\\
\includegraphics[height=2.3cm]{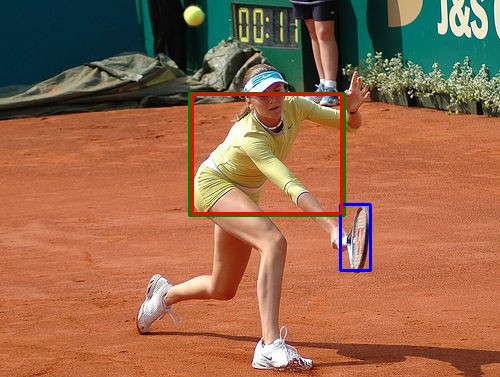} & \includegraphics[height=2.3cm]{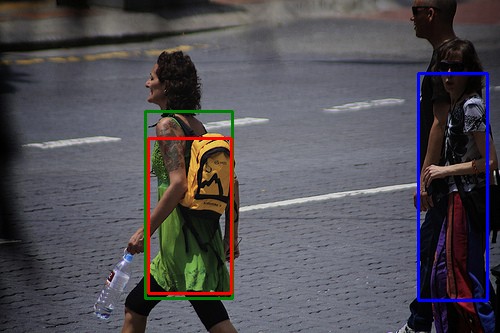} & \includegraphics[height=2.3cm]{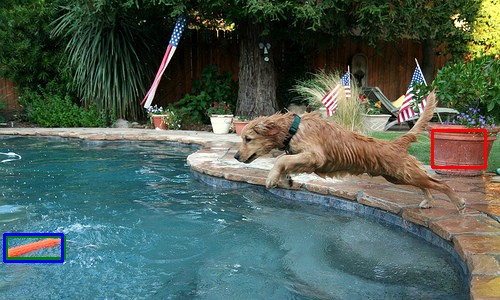} & \includegraphics[height=2.3cm]{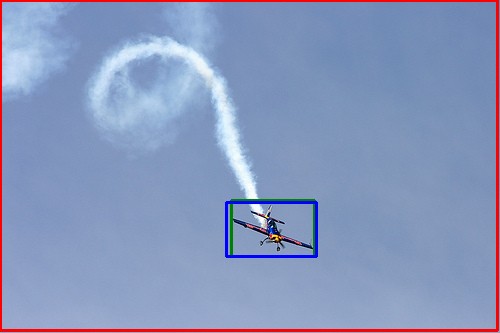}\\
    \noalign{\smallskip}
sky & pink blanket & peeps & guy in yellow shirt\\
\includegraphics[height=2.3cm]{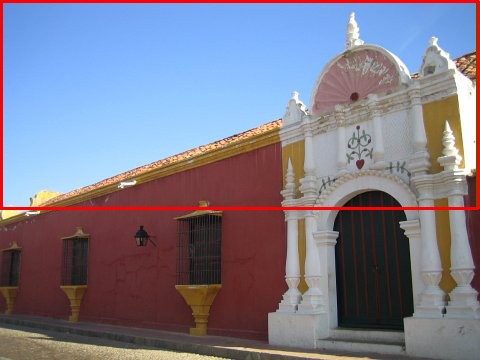} & \includegraphics[height=2.3cm]{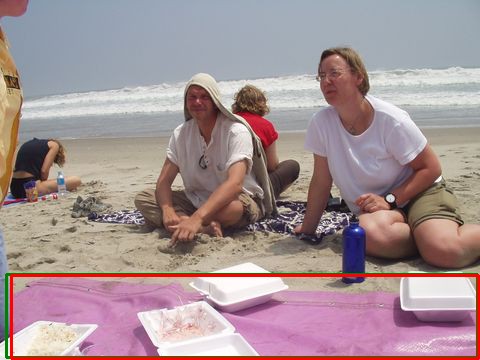} & \includegraphics[height=2.3cm]{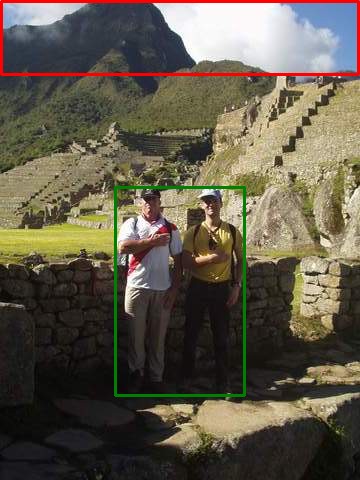} & \includegraphics[height=2.3cm]{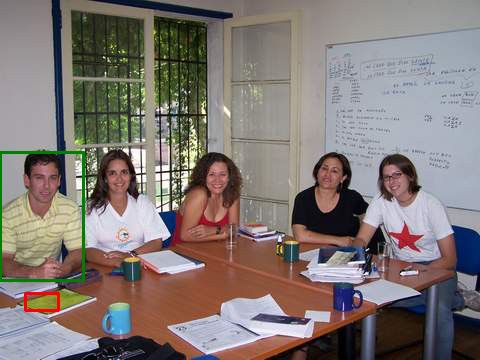}\\  
    \end{tabular}
   \caption{Example localization output for Flickr30kEntities (top row) and ReferItGame (bottom row). We compare the effects of adding a \textbf{colour} detector (\textcolor{darkred}{red} bounding box) to \textbf{tfcoco+tfoid+places365} (\textcolor{darkblue}{blue}) (\textbf{w2v-avg}, \textbf{union}). The ground truth is indicated in \textcolor{darkgreen}{green}. The first two columns show examples of where adding a \textbf{colour} detector improves localization, while the last two columns are examples where it hurts localization.}
\label{fig:output-colours}
\end{figure*}

\begin{comment}
\begin{figure*}[!t]
  \centering
  \begin{tabular}{cccc}
a white dog & white shorts & cats & blue\\
\includegraphics[height=2.4cm]{3482974845-0-121622} & \includegraphics[height=2.4cm]{2763601657-0-71291} & \includegraphics[height=2.4cm]{3539817989-4-125987} & \includegraphics[height=2.4cm]{4818675994-2-206758}\\
    \noalign{\smallskip}
red book & yellow chair & skyscraper & blue cloth\\
\includegraphics[height=2.4cm]{27261-27261-9} & \includegraphics[height=2.4cm]{4640-4640-3} & \includegraphics[height=2.4cm]{40323-40323-1}
 & \includegraphics[height=2.4cm]{18694-18694-7}\\
    \end{tabular}
   \caption{Example localization output for Flickr30kEntities (top row) and ReferItGame (bottom row). We compare the effects of using a \textbf{consensus} (\textcolor{darkred}{red} bounding box) against a \textbf{union} (\textcolor{darkblue}{blue}) localization strategy (with \textbf{tfcoco+tfoid+places365+color}, \textbf{w2v-avg}). The ground truth is indicated in \textcolor{darkgreen}{green}. The first two columns show examples of where using \textbf{consensus} improves localization, while the last two columns are examples where it hurts localization.}
\label{fig:output-consensus}
\end{figure*}
\end{comment}

%

\paragraph{Upperbound.} The upperbound accuracies generally increase as we increase the number of detectors and categories used. This indicates that the %
recall has increased, presumably by virtue of more candidate bounding boxes being proposed. Our concept selection process 
reduces this upperbound, but as a result allows the localization strategy to perform the task more accurately. Interestingly, the upperbound did not change significantly when only a subset of $20$ categories %
of \textbf{tfcoco} is used and with concept selection applied ($51.81\%$ vs.\ $51.22\%$ in Table~\ref{tbl:results-flickr30k}). This is because of the large number of people-related phrases in the dataset; the \emph{person} detectors in both detector groups manage to capture this. 

\vspace*{-12pt}
\paragraph{Detector.} The accuracy generally improves with more detectors (and number of categories), as long as the %
detections are of high quality. While the differences %
between \textbf{tfcoco20} ($20$ categories) and \textbf{tfcoco} ($80$ categories) are much smaller, using \textbf{tfoid} ($545$ categories) resulted in larger improvements (see Figure~\ref{fig:output-oid}). %
Detector quality is also important, as demonstrated by the generally weak performance from \textbf{yolo9000} (accuracies are generally lower than $20\%$) which has a low accuracy ($19.7$ mAP on a subset of $200$ categories \cite{RedmonFarhadi:2017}) despite boasting the ability to detect over $9,000$ categories. The category labels themselves include abstract categories (\emph{thing}, \emph{instrumentation}), which are irrelevant as these are not often used to describe objects. The \textbf{colour} detectors on their own gave low accuracies (generally \textless $10\%$). This is because only a small subset of test phrases contained colour terms, and the connected component labelling also resulted in generally small bounding boxes; however, using the union of bounding boxes for localization resulted in better accuracies (${\approx}18\%$).

\vspace*{-12pt}
\paragraph{Combination of detectors.} The detectors are also complementary to each other. Combining \textbf{tfcoco} and \textbf{tfoid} results in a higher accuracy ($48.20\%$) than using either alone ($37.57\%$ and $44.69\%$ respectively). From Table~\ref{tbl:catresults-flickr30k}, we observe that \textbf{tfoid} helped improve over \textbf{tfcoco} especially for \emph{clothing} (by ${\approx}27\%$ accuracy), \emph{bodyparts} (${\approx}20\%$) and \emph{instruments} (${\approx}30\%$), and to a certain extent \emph{scenes}. It also provided some additional redundancy to help %
localize \emph{person} since it contains different people detectors (\emph{person}, \emph{man}, \emph{woman}, \emph{boy}, \emph{girl}). \textbf{places365} improved the localization of scene phrases (${\approx}19\%$).

\vspace*{-12pt}
\paragraph{Colour detector.} Adding a \textbf{colour} detector to \textbf{tfcoco+tfoid+places365} does not improve the overall accuracies (CC+OI+PL (avg, u) vs.\ CC+OI+PL+CL (avg, u) in Table~\ref{tbl:catresults-flickr30k}), but it helps with scene-type phrases, especially when the scene contains a colour term and covers most of the image. %
It also helps %
when the phrase is a single colour noun (\emph{red}), when the head noun is not detected (\emph{an orange outfit}), or when the colour can be inferred for a missed detection (\emph{tree}). This works as long as there are no other objects with the same colour. Some problematic cases are when the desired colour occurs elsewhere in the image, and with phrases such as \emph{a white man}. 
Figure~\ref{fig:output-colours} shows some examples illustrating the contributions of the \textbf{colour} detector. We further quantitatively investigate the contributions of \textbf{colour} to \textbf{tfcoco+tfoid+places365} by evaluating on a subset of test phrases where the 11 basic colour terms occur (Table~\ref{tbl:catresults-flickr30k-color}). %
We observe that colour terms are most frequently mentioned in \emph{clothing}-type phrases. Adding a \textbf{colour} detector improves localization in \emph{clothing} and \emph{scene} phrases.

\begin{table*}[t]
\centering
\resizebox{\linewidth}{!}{
\small
\begin{tabular}{lcccccccc|c}
\toprule
 & \textbf{people} & \textbf{clothing} & \textbf{bodyparts} & \textbf{animals} & \textbf{vehicles} & \textbf{instruments} & \textbf{scene} 
& \textbf{other} & \textbf{overall} \\
\midrule
\# instances & 30 & 1323 & 48 & 177 & 79 & 1 & 93 & 292 & 2033 \\
\midrule
CC+OI+PL (avg, u) & 66.67 & 34.24 & 29.17 & 89.27 & 79.75 & 100.00 & 44.09 & 48.63 & 43.43 \\
CC+OI+PL+CL (avg, u) & 40.00 & 35.75 & 27.08 & 69.49 & 65.82 & 100.00 & 58.06 & 44.86 & 41.81 \\
\bottomrule
\end{tabular}
}
\caption{Phrase localization accuracies for different phrase types on a subset of query phrases that contain at least one basic colour term.}
\label{tbl:catresults-flickr30k-color}
\end{table*}

\vspace*{-12pt}
\paragraph{Concept selection.} %
Our concept selection process with word embeddings %
similarity is intuitive, and %
results in accurate localization. \textbf{tfcoco} performed on par with Yeh~\etal~\cite{YehEtAl:2018} which computes the similarity using paired annotations, %
and with the same $80$ categories. Our approach captures distributional similarities, and is undesirable in certain cases, for example a \emph{cyclist} is more similar to a \emph{bicycle} or a \emph{wheel} than it is to a \emph{person}. We also found that all three word vector aggregation schemes perform comparably; \textbf{w2v-last} generally performs similar to \textbf{w2v-max} with only a minor degradation. This agrees with our assumption that the last word in the phrase is most likely the head word in Flickr30kEntities. \textbf{w2v-avg} also performs very slightly worse in general, %
except when \textbf{colour} detectors are used.

\vspace*{-12pt}
\paragraph{Localization strategy.} For this dataset, the \textbf{union} localization strategy seems to work best, partly because of how the test dataset is constructed. %
It is also useful for \textbf{colour} detectors which produce generally small bounding boxes. The \textbf{largest} strategy also works reasonably well; objects mentioned in the captions tend to be larger than those not. %
Our novel \textbf{consensus} strategy, designed to allow for slightly higher upperbounds and %
accuracies by voting, generally gave accuracies comparable to \textbf{union}-based equivalents. %

\subsection{Phrase localization on ReferItGame}
\label{sec:results-refclef}

\textbf{ReferItGame}~\cite{KazemzadehEtAl:2014} crowdsources phrases from an interactive game to describe segments in %
IAPR TC-12 images~\cite{GrubingerEtAl:2006}. It is significantly different from Flickr30kEntities as phrases are not extracted from image captions and are also much shorter.
We use the test split of Rohrbach~\etal~\cite{RohrbachEtAl:2016} consisting of $65,193$ phrases for $9,999$ images\footnote{We used the split provided at \url{https://github.com/lichengunc/refer}}. Again, %
the training and validation splits are ignored. %

\begin{table}[t]
  \centering
  	\small
  	\begin{tabular}{l l c l}
    \toprule
    \textbf{Detector} & \textbf{Similarity} & \textbf{Strategy} & \textbf{Acc (UB) \%}\\
	\midrule
    \multicolumn{3}{l}{Baseline: Always localize to whole image} & 14.64\\
    20 & w2v-max & largest & 14.97 (26.82)\\
    CC & w2v-max & largest & 15.40 (27.16)\\
    OI & w2v-max & largest & 19.82 (28.03)\\
    CC+OI & w2v-avg & largest & 21.21 (32.70)\\	    
    CC+OI+PL & w2v-avg & consensus & 22.25 (35.56)\\
    CC+OI+PL & w2v-avg & largest & 23.95 (35.04)\\ 
    CC+OI+PL+CL & w2v-max & consensus & 25.52 (42.48)\\
    CC+OI+PL+CL & w2v-max & largest & \textbf{\underline{26.48}} (39.50)\\ 
    \midrule
    \multicolumn{4}{l}{\textbf{\emph{Weakly supervised}}}\\
    GroundeR~\cite{RohrbachEtAl:2016} & & & 10.70 \\    
    \multicolumn{2}{l}{KAC Net + Soft KBP~\cite{ChenEtAl:2018}} & & 15.83 \\    
    Yeh~\etal~\cite{YehEtAl:2018} & & yolococo & 17.96\\
    Yeh~\etal~\cite{YehEtAl:2018} & \multicolumn{2}{r}{vgg+yolococo} & 20.91\\ %
    \noalign{\medskip}
    \multicolumn{4}{l}{\textbf{\emph{Strongly supervised}}}\\
    GroundeR~\cite{RohrbachEtAl:2016} & & & 26.93 \\       
    Hu~\etal~\cite{HuEtAl:2016} & & & 27.80 \\
    QRC Net~\cite{ChenEtAl:2017} & & & 44.07\\      
    \bottomrule
    \end{tabular}
\caption{Accuracies of some of our selected models on ReferItGame. Again, we present our results alongside selected strongly and weakly supervised systems as a comparison since no previous non-paired model exists. Keys: CC=\textbf{tfcoco}, OI=\textbf{tfoid}, 20=\textbf{tfcoco20}, PL=\textbf{places365}, CL=\textbf{colour}.}
\label{tbl:results-refclef}
\end{table}

Table~\ref{tbl:results-refclef} shows the accuracies on ReferItGame for a selected set of our models, again with different combinations of detectors, concept selection and localization strategies. 
The accuracy of the baseline of always localizing to the whole image is $14.64\%$. Our best performing model again performs better than all weakly supervised models ($26.48\%$ vs.\ the state of the art's $20.91\%$), and is on par with some strongly supervised models~\cite{HuEtAl:2016,RohrbachEtAl:2016}, although not at the level of QRC Net~\cite{ChenEtAl:2017}. %

\subsubsection{Discussion}

\paragraph{Localization strategy.} Unlike Flickr30kEntities, taking the \textbf{union} does not perform as well as simply taking the \textbf{largest} box; this is consistent across models. Again, our proposed \textbf{consensus} strategy performs well, although not generally as well as the \textbf{largest} strategy.

\vspace*{-12pt}
\paragraph{Concept selection.} Like Flickr30kEntities, \textbf{w2v-max} and \textbf{w2v-avg} performs equally well, with \textbf{w2v-max} having a very slight edge. Unlike Flickr30kEntities, \textbf{w2v-last} performs substantially worse than other semantic similarity measures. This is because the phrases are short, and the head word is more likely to be mentioned at the beginning. %

\vspace*{-12pt}
\paragraph{Detectors.} The detectors %
generally show a similar behaviour as with Flickr30kEntities. Adding \textbf{tfoid} to \textbf{tfcoco} pushed the accuracy beyond the state of the art~\cite{YehEtAl:2018}, and adding \textbf{places365} further increased its accuracy. Unlike Flickr30kEntities, the \textbf{colour} detector contributed substantially more, %
increasing the overall accuracy by ${\approx}3\%$. %
ReferItGame has many colour-based phrases (many single colour words), due to how the annotations were obtained. The model also performs well by inferring the colour of \emph{sky}, \emph{cloud} and \emph{tree}, which occur frequently %
(Figure~\ref{fig:output-colours}). 

\subsection{Phrase localization on Visual Genome}
\label{sec:results-visgenome}

To demonstrate our model's scalability to different datasets, we also test our model on Visual Genome~\cite{KrishnaEtAl:2017} where the queries are at sentence level, rather than at phrase level. %
Zhang~\etal~\cite{ZhangEtAl:2017} reported $26.4\%$
localization accuracy with a fully supervised method. The only weakly supervised equivalent of which we are aware reported $24.4\%$ accuracy~\cite{XiaoEtAl:2017}, but this is an unfair comparison because they evaluated whether %
a single point falls inside the bounding box, rather than predicting the full box. Our model (\textbf{tfoid}, \textbf{w2v-max}, \textbf{largest}) achieved $14.29\%$ accuracy on Visual Genome, and by combining \textbf{tfoid} with \textbf{tfcoco}  the accuracy is increased to $16.39\%$. This observation is consistent to what we reported, %
and we infer that the same pattern should apply to further combinations and variants.

\section{Conclusions}

We introduced the first approach to phrase localization in images without %
phrase localization annotations. 
This non-paired approach, while simple, proved %
effective: In experiments with %
Flickr30kEntities and ReferItGame %
it outperformed all existing weakly supervised approaches and performed %
competitively to strongly supervised approaches. The method is a strong baseline -- phrase localization can be successfully performed on these datasets even without paired examples. Our work suggests that there is significant room for simpler and general methods that rely on few/no paired annotations, instead of complex models that attempt to fit paired annotations to achieve high performance improvements without the ability to generalize. %
This finding can change how Language \& Vision tasks are viewed and tackled in future -- researchers should make better use of paired annotations beyond what can already be achieved without such task-specific data.

{\small
\bibliographystyle{ieee_fullname}
\bibliography{iccv19}
}
\section*{Appendix}
\label{sec:appendix}

\begin{figure*}[hbt]
  \begin{center}
    \includegraphics[width=1.0\linewidth]{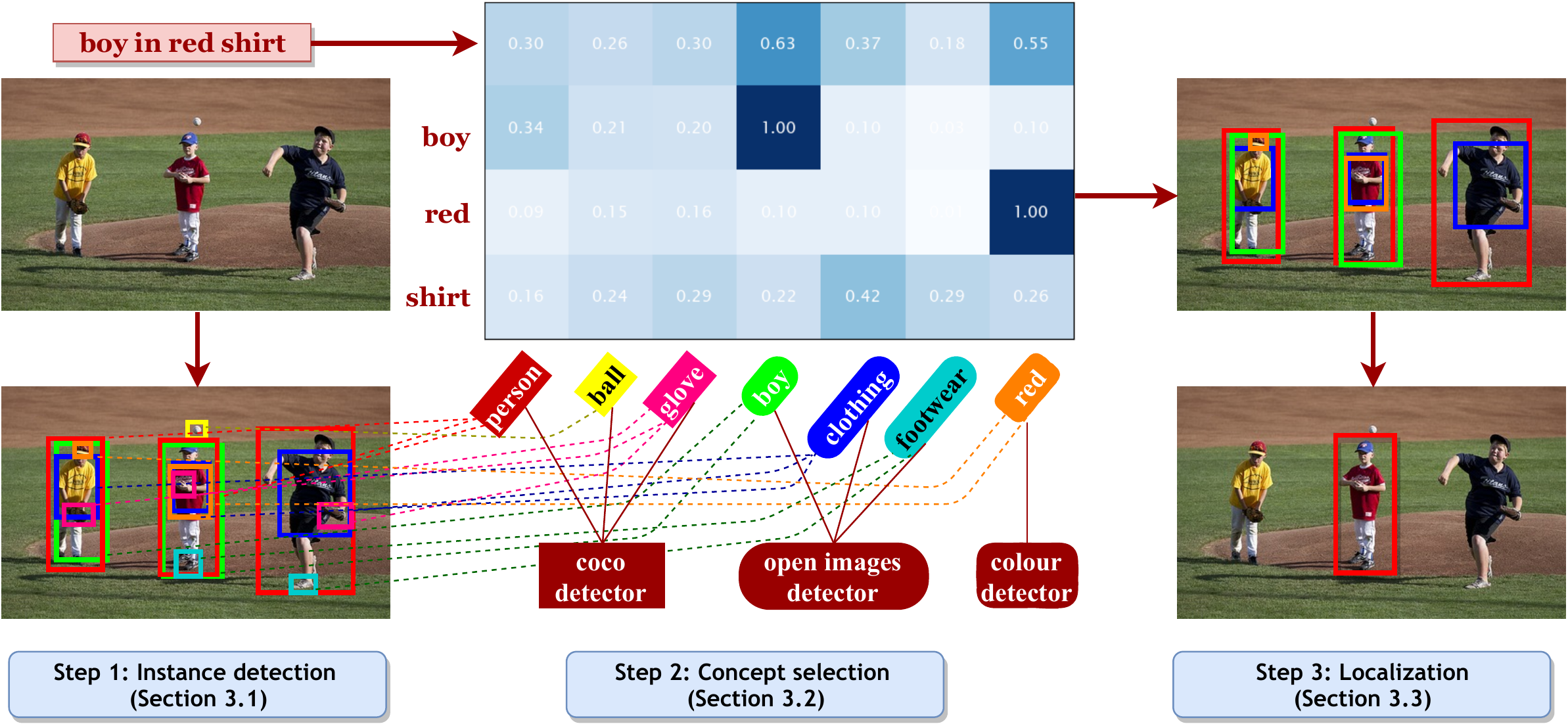}
  \end{center}
   \caption{The three stages of the proposed model for unsupervised phrase localization. The \textbf{instance detection} phase detects instances of various concepts using pre-trained detectors. The \textbf{concept selection} stage ranks these detected concepts against the query phrase (using pre-trained word embeddings) and forwards the best candidate concept instance(s) to the \textbf{localization} phase, where the model predicts the final bounding box for the query phrase.}
\label{fig:model-full}
\end{figure*}

Figure~\ref{fig:model-full} gives a full-sized illustration of our proposed model for phrase localization without paired examples.\\

Table~\ref{tbl:fullaccuracy-flickr30k} shows the complete phrase localization accuracies on the Flickr30kEntities dataset, for different detector combinations, semantic similarities and localization strategies. The upper-bounds for each are also provided (the best possible achievable accuracy before performing the phrase localization step). The upper-bound variant \textbf{+union} includes the minimal bounding box encompassing all candidates as a candidate bounding box, while \textbf{-union} does not. Note that the \textbf{places365} detector is equivalent to a baseline that always localizes a phrase to the whole image.\\

Table~\ref{tbl:fullaccuracy-refclef} shows the complete accuracies on the ReferItGame test dataset, again for different combinations of detectors, concept selection measure and localization strategies.\\

\begin{figure*}[!t]
  \begin{center}
    \begin{tabular}{cccc}
skyscrapers & a blue swimsuit & the Arc de Triomphe & bright red beanie\\
\includegraphics[height=2.4cm]{3707283973-0-138377} & \includegraphics[height=2.4cm]{2823575468-2-74927} & \includegraphics[height=2.4cm]{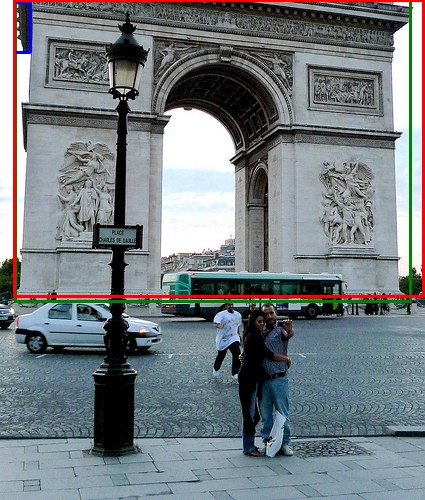} & \includegraphics[height=2.4cm]{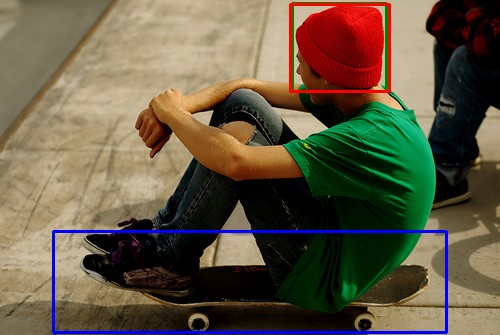}\\
three men & A very excited drummer & Women & A five member band\\
\includegraphics[height=2.4cm]{44619131-1-170436} & \includegraphics[height=2.4cm]{4803926618-4-204441} & \includegraphics[height=2.4cm]{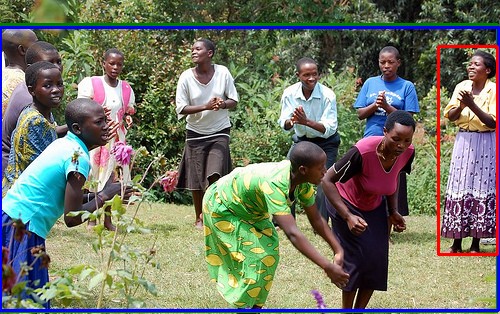} & \includegraphics[height=2.4cm]{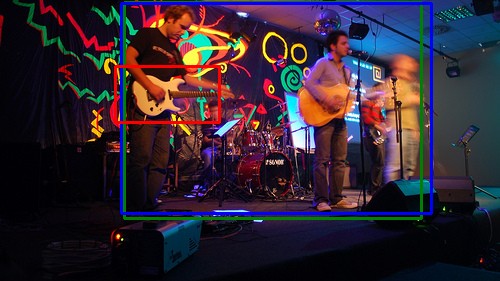}\\
    \end{tabular}
  \end{center}
   \caption{Example localization output for Flickr30kEntities. We compare the effects of adding a \textbf{tfoid} detector (\textcolor{darkred}{red} bounding box) to \textbf{tfcoco} (\textcolor{darkblue}{blue}) (\textbf{w2v-max}, \textbf{union}). The ground truth is indicated in \textcolor{darkgreen}{green}. The first row shows examples of where adding a \textbf{tfoid} detector improves localization, while the last row provides examples where it hurts localization.}
\label{fig:output-oid-flickr30k}
\end{figure*}

\begin{figure*}[!b]
  \begin{center}
    \begin{tabular}{cccc}
sky & lamp & hotel door & front right tire\\
\includegraphics[height=2.4cm]{38163-38163-5} & \includegraphics[height=2.4cm]{12290-12290-5} & \includegraphics[height=2.4cm]{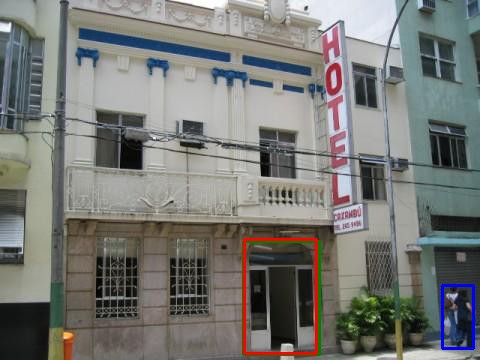}
 & \includegraphics[height=2.4cm]{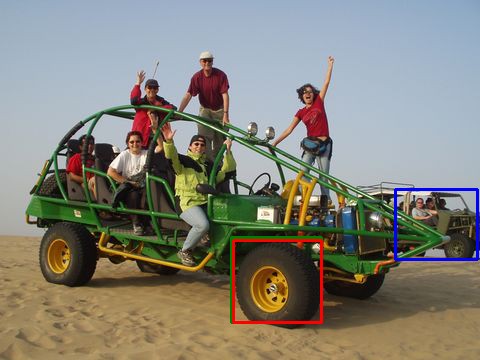}\\
soda & glass & fishy on right & pick his nose\\
\includegraphics[height=2.4cm]{7002-7002-11} & \includegraphics[height=2.4cm]{9403-9403-1} & \includegraphics[height=2.4cm]{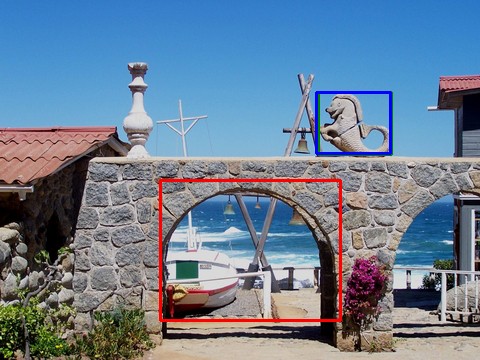} 
& \includegraphics[height=2.4cm]{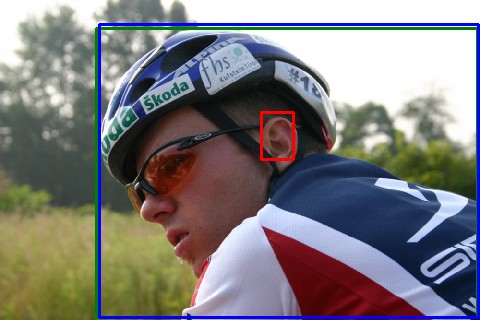}\\
    \end{tabular}
  \end{center}
   \caption{Example localization output for ReferItGame. We compare the effects of adding a \textbf{tfoid} detector (\textcolor{darkred}{red} bounding box) to \textbf{tfcoco} (\textcolor{darkblue}{blue}) (\textbf{w2v-max}, \textbf{union}). The ground truth is indicated in \textcolor{darkgreen}{green}. The first row shows examples of where adding a \textbf{tfoid} detector improves localization, while the last row provides examples where it hurts localization.}
\label{fig:output-oid-refclef}
\end{figure*}

Figures~\ref{fig:output-oid-flickr30k} and \ref{fig:output-oid-refclef} show some example output from our model on Flickr30kEntities and ReferItGame respectively, where we compare the effects of combining a \textbf{tfoid} detector with a \textbf{tfcoco} detector.\\

\begin{figure*}[th]
  \begin{center}
    \begin{tabular}{cccc}
a yellow tennis suit & a long green shirt & a red slide & a purple shirt\\
\includegraphics[height=2.4cm]{23016347-0-40122} & \includegraphics[height=2.4cm]{4623878206-0-183941} & \includegraphics[height=2.4cm]{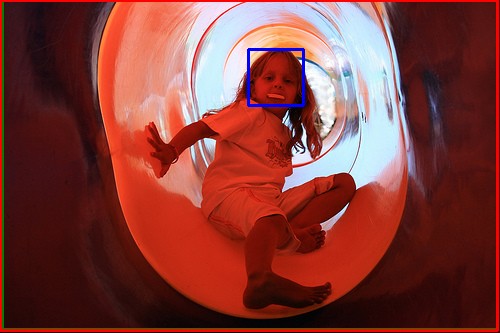} & \includegraphics[height=2.4cm]{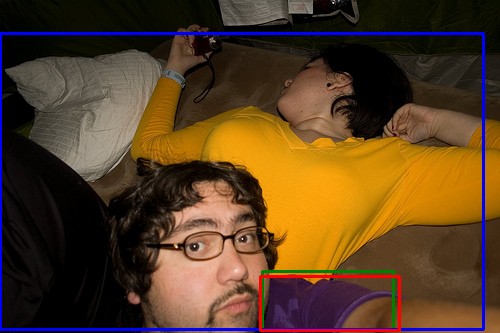}\\
a red toy & A blue , red , and yellow airplane & red , white , and blue & Older male\\
\includegraphics[height=2.4cm]{1288909046-1-7434} & \includegraphics[height=2.4cm]{3584603849-2-129624} & \includegraphics[height=2.4cm]{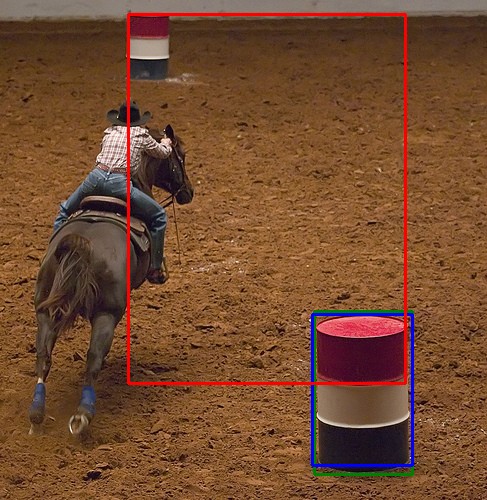} & \includegraphics[height=2.4cm]{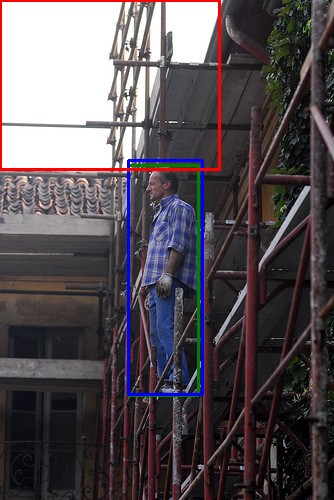}\\
    \end{tabular}
  \end{center}
   \caption{Example localization output for Flickr30kEntities. We compare the effects of adding a \textbf{colour} detector (\textcolor{darkred}{red} bounding box) to \textbf{tfcoco+tfoid+places365} (\textcolor{darkblue}{blue}) (\textbf{w2v-avg}, \textbf{union}). The ground truth is indicated in \textcolor{darkgreen}{green}. The first row shows examples of where adding a \textbf{colour} detector improves localization, while the last row provides examples where it hurts localization.}
\label{fig:output-colours-flickr30k}
\end{figure*}

\begin{figure*}[th]
  \begin{center}
    \begin{tabular}{cccc}
sky & pink blanket & trees & blue shirt\\
\includegraphics[height=2.4cm]{27357-27357-2} & \includegraphics[height=2.4cm]{18837-18837-3} & \includegraphics[height=2.4cm]{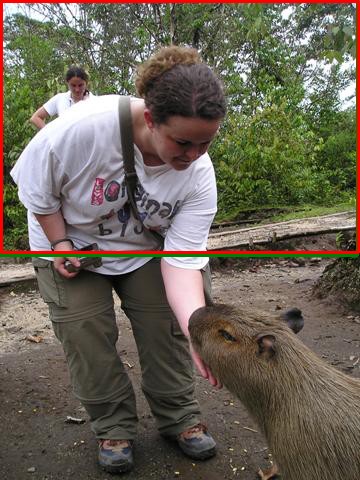} & \includegraphics[height=2.4cm]{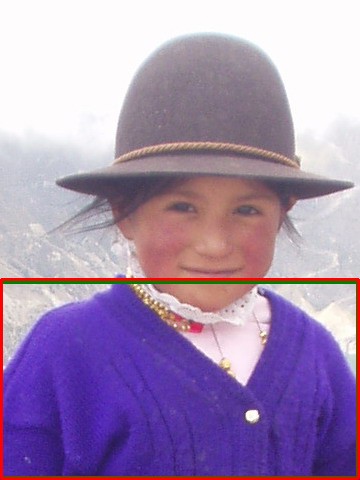}\\
peeps & guy in yellow shirt & the paper & bag below women in orange\\
\includegraphics[height=2.4cm]{23977-23977-1} & \includegraphics[height=2.4cm]{23592-23592-4} & \includegraphics[height=2.4cm]{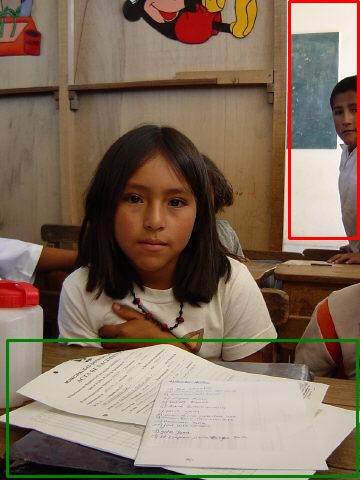} &
 \includegraphics[height=2.4cm]{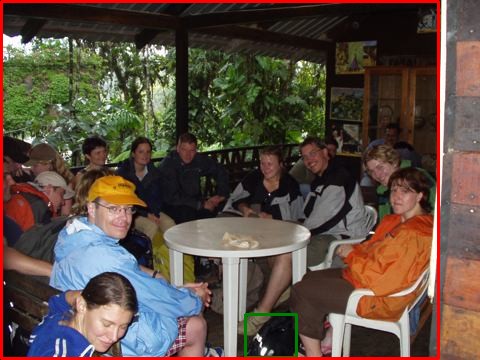}\\
    \end{tabular}
  \end{center}
   \caption{Example localization output for ReferItGame. We compare the effects of adding a \textbf{colour} detector (\textcolor{darkred}{red} bounding box) to \textbf{tfcoco+tfoid+places365} (\textcolor{darkblue}{blue}) (\textbf{w2v-avg}, \textbf{union}). The ground truth is indicated in \textcolor{darkgreen}{green}. The first row shows examples of where adding a \textbf{colour} detector improves localization, while the last row provides examples where it hurts localization.}
\label{fig:output-colours-refclef}
\end{figure*}

Figures~\ref{fig:output-colours-flickr30k} and \ref{fig:output-colours-refclef} show some example output from our model on Flickr30kEntities and ReferItGame respectively, where we compare the effects of combining a \textbf{colour} detector with a \textbf{tfcoco+tfoid+places365} detector.\\

\begin{figure*}[th]
  \begin{center}
    \begin{tabular}{cccc}
a white dog & white shorts & brown pants & a glass cup\\
\includegraphics[height=2.4cm]{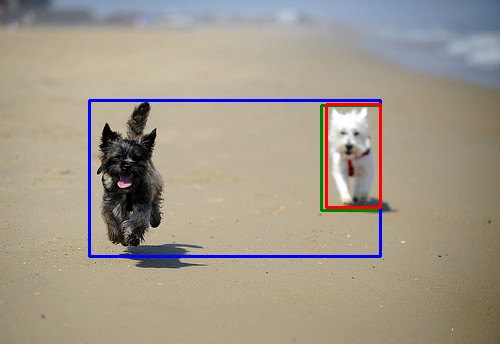} & \includegraphics[height=2.4cm]{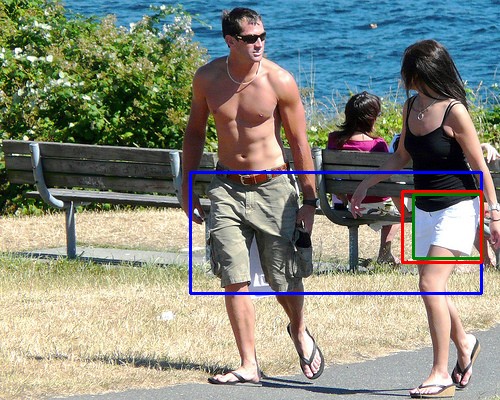} & \includegraphics[height=2.4cm]{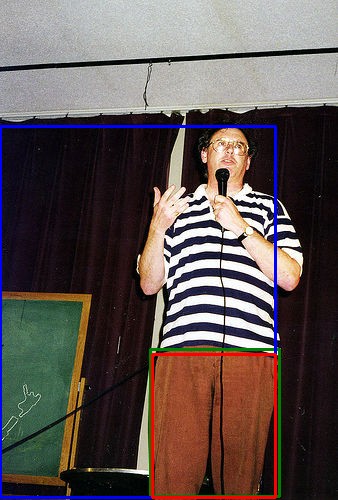} & \includegraphics[height=2.4cm]{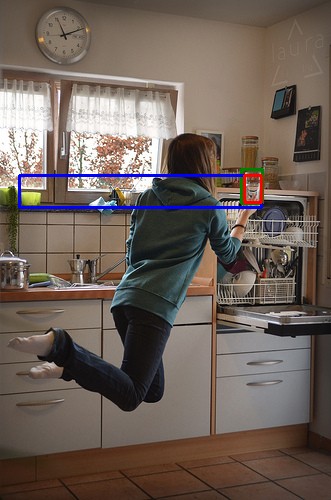}\\
cats & blue & a blue and white house & red , white , and green colored costumes\\
\includegraphics[height=2.4cm]{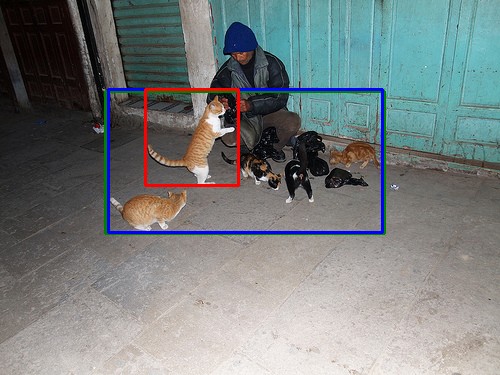} & \includegraphics[height=2.4cm]{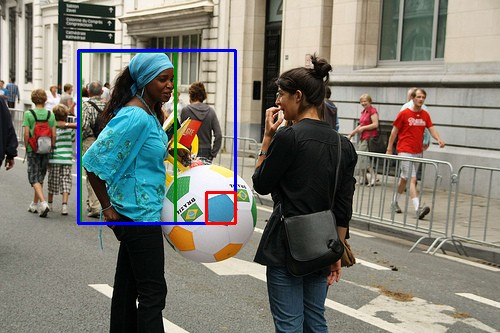} & \includegraphics[height=2.4cm]{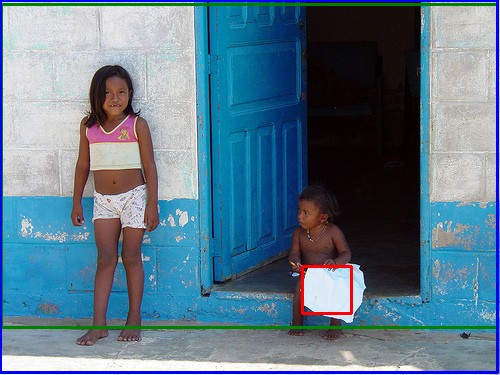} & \includegraphics[height=2.4cm]{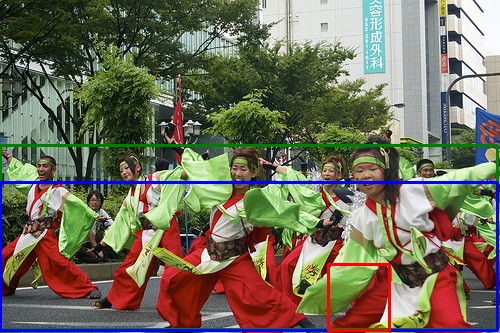}\\
    \end{tabular}
  \end{center}
   \caption{Example localization output for Flickr30kEntities. We compare the effects of using a \textbf{consensus} (\textcolor{darkred}{red} bounding box) against a \textbf{union} (\textcolor{darkblue}{blue}) localization strategy (with \textbf{tfcoco+tfoid+places365+color}, \textbf{w2v-avg}). The ground truth is indicated in \textcolor{darkgreen}{green}. The first row shows examples of where using \textbf{consensus} improves localization, while the last row provides examples where it hurts localization.}
\label{fig:output-consensus-flickr30k}
\end{figure*}

\begin{figure*}[!t]
  \begin{center}
    \begin{tabular}{cccc}
red book & yellow chair & black truck on rode next to motorcycle & yellow / orange thing right of boy\\
\includegraphics[height=2.4cm]{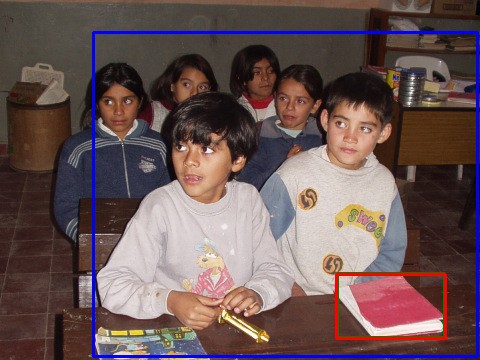} & \includegraphics[height=2.4cm]{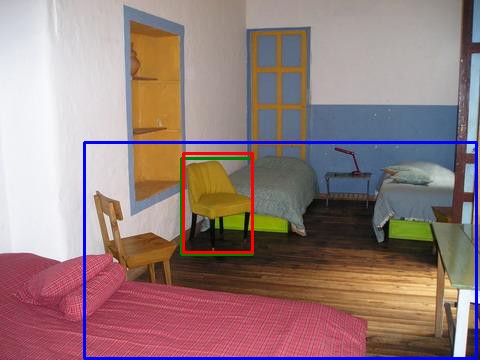} & \includegraphics[height=2.4cm]{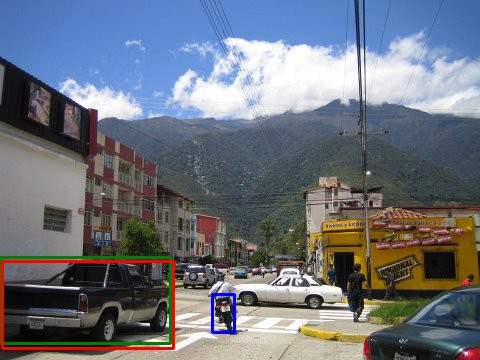} & \includegraphics[height=2.4cm]{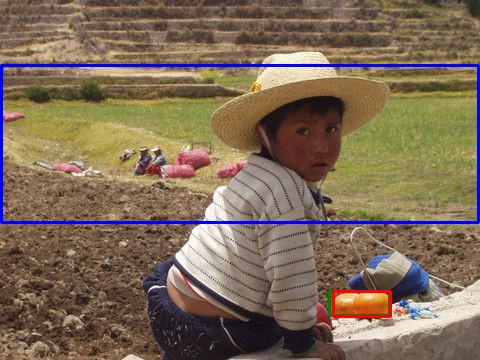}\\
skyscraper & blue cloth & red & people\\
\includegraphics[height=2.4cm]{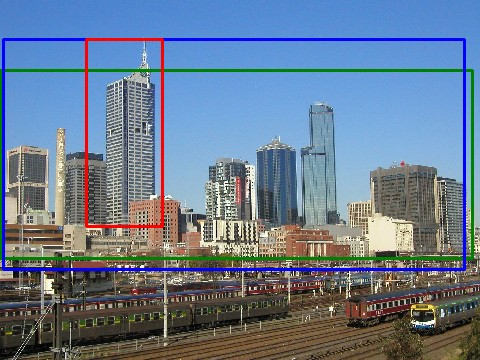} & \includegraphics[height=2.4cm]{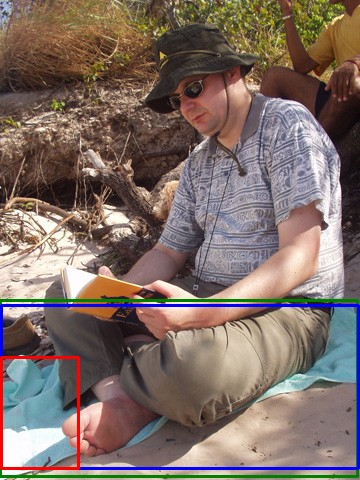} & \includegraphics[height=2.4cm]{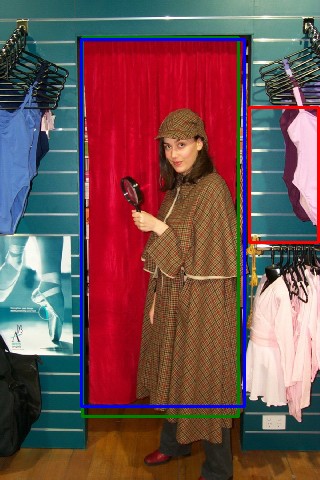} & \includegraphics[height=2.4cm]{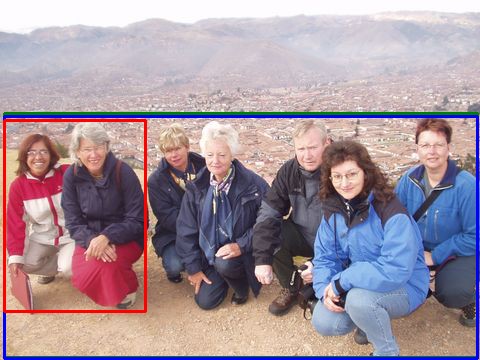}\\
    \end{tabular}
  \end{center}
   \caption{Example localization output for ReferItGame. We compare the effects of using a \textbf{consensus} (\textcolor{darkred}{red} bounding box) against a \textbf{union} (\textcolor{darkblue}{blue}) localization strategy (with \textbf{tfcoco+tfoid+places365+color}, \textbf{w2v-avg}). The ground truth is indicated in \textcolor{darkgreen}{green}. The first row shows examples of where using \textbf{consensus} improves localization, while the last row provides examples where it hurts localization.}
\label{fig:output-consensus-refclef}
\end{figure*}

Figures~\ref{fig:output-consensus-flickr30k} and \ref{fig:output-consensus-refclef} show some example output from our model on Flickr30kEntities and ReferItGame respectively, where we compare the effects of using our proposed \textbf{consensus} localization strategy against a \textbf{union} strategy.

\newpage

\begin{table*}
  \begin{center}
  	\small
  	\begin{tabular}{l l |cc| ccccc}
    \toprule
    \multirow{2}{*}{\textbf{Detector}} & \multirow{2}{*}{\textbf{Similarity}} & \multicolumn{2}{c|}{\textbf{Upperbound}} & \multicolumn{5}{c}{\textbf{Localization strategy}} \\
        & & -union & +union & random & largest & confidence & union & consensus\\
\midrule
\multirow{4}{*}{\parbox{2.75cm}{tfcoco20}} & no filter & 51.67 & 58.83 & 14.62 & 32.89 & 30.23 & 30.93 & -\\
 & w2v-max & 45.28 & 51.54 & 22.21 & 35.98 & 32.62 & 36.33 & -\\
 & w2v-last & 44.92 & 51.16 & 21.52 & 35.72 & 32.40 & 36.13 & -\\
 & w2v-avg & 45.49 & 51.81 & 22.26 & 36.13 & 32.79 & 36.49 & -\\
\midrule
\multirow{4}{*}{\parbox{2.75cm}{tfcoco}} & no filter & 58.22 & 65.31 & 11.37 & 32.53 & 30.08 & 29.81 & -\\
 & w2v-max & 45.32 & 51.22 & 24.10 & 37.03 & 34.20 & 37.57 & -\\
 & w2v-last & 45.30 & 51.21 & 24.51 & 37.03 & 34.18 & 37.70 & -\\
 & w2v-avg & 45.17 & 51.29 & 24.06 & 37.03 & 34.11 & 37.81 & -\\
\midrule
\multirow{4}{*}{\parbox{2.75cm}{tfoid}} & no filter & 54.26 & 63.26 & 17.28 & 28.95 & 21.86 & 30.90 & -\\
 & w2v-max & 43.83 & 50.04 & 38.27 & 40.59 & 39.77 & 44.69 & -\\
 & w2v-last & 43.69 & 49.98 & 38.30 & 40.47 & 39.69 & 44.54 & -\\
 & w2v-avg & 43.24 & 49.56 & 37.86 & 40.04 & 39.22 & 44.31 & -\\
\midrule
\multirow{4}{*}{\parbox{2.75cm}{yolo9000}} & no filter & 24.94 & 36.25 & 6.97 & 18.20 & 11.82 & 22.37 & -\\
 & w2v-max & 15.14 & 20.85 & 11.82 & 13.74 & 12.51 & 17.79 & -\\
 & w2v-last & 15.10 & 20.69 & 11.53 & 13.76 & 12.57 & 17.72 & -\\
 & w2v-avg & 15.14 & 20.99 & 11.50 & 13.76 & 12.53 & 17.89 & -\\
\midrule
\multirow{4}{*}{\parbox{2.75cm}{places365}} & no filter & 21.99 & 21.99 & 21.99 & 21.99 & 21.99 & 21.99 & -\\
 & w2v-max & 21.99 & 21.99 & 21.99 & 21.99 & 21.99 & 21.99 & -\\
 & w2v-last & 21.99 & 21.99 & 21.99 & 21.99 & 21.99 & 21.99 & -\\
 & w2v-avg & 21.99 & 21.99 & 21.99 & 21.99 & 21.99 & 21.99 & -\\
\midrule
\multirow{4}{*}{\parbox{2.75cm}{color}} & no filter & 31.71 & 43.64 & 1.06 & 13.91 & 4.32 & 22.26 & -\\
 & w2v-max & 10.19 & 23.11 & 3.67 & 8.51 & 5.13 & 18.51 & -\\
 & w2v-last & 6.53 & 18.67 & 2.23 & 5.50 & 2.88 & 15.87 & -\\
 & w2v-avg & 10.26 & 23.23 & 3.85 & 8.62 & 5.19 & 18.65 & -\\
\midrule
\multirow{4}{*}{\parbox{2.75cm}{tfcoco+tfoid}} & no filter & 67.24 & 73.00 & 13.39 & 30.32 & 30.32 & 27.47 & -\\
 & w2v-max & 48.84 & 55.85 & 38.62 & 44.00 & 41.95 & 48.20 & 47.91\\
 & w2v-last & 48.70 & 55.80 & 38.63 & 43.92 & 41.83 & 48.14 & -\\
 & w2v-avg & 48.31 & 55.37 & 38.21 & 43.53 & 41.44 & 47.86 & 47.63\\
\midrule
\multirow{4}{*}{\parbox{2.75cm}{tfcoco+tfoid+ yolo9000}} & no filter & 67.98 & 73.20 & 11.77 & 28.42 & 30.32 & 23.09 & -\\
 & w2v-max & 46.68 & 53.34 & 37.38 & 42.41 & 40.70 & 46.36 & 46.03\\
 & w2v-last & 46.76 & 53.44 & 37.43 & 42.55 & 40.81 & 46.36 & -\\
 & w2v-avg & 45.22 & 51.99 & 36.58 & 41.21 & 39.74 & 45.37 & 45.22\\
\midrule
\multirow{4}{*}{\parbox{2.75cm}{tfcoco+tfoid+ places365}} & no filter & 72.63 & 73.37 & 13.85 & 22.97 & 30.34 & 22.73 & -\\
 & w2v-max & 51.56 & 57.81 & 41.70 & 46.96 & 45.00 & 50.49 & 50.29\\
 & w2v-last & 51.45 & 57.81 & 41.78 & 46.87 & 44.90 & 50.46 & -\\
 & w2v-avg & 51.21 & 57.51 & 41.70 & 46.63 & 44.71 & 50.27 & 50.11\\
\midrule
\multirow{4}{*}{\parbox{2.75cm}{tfcoco+tfoid+ places365+yolo9000}} & no filter & 73.31 & 73.99 & 12.01 & 22.79 & 30.34 & 19.12 & -\\
 & w2v-max & 49.52 & 55.59 & 40.71 & 45.35 & 43.74 & 48.73 & 48.51\\
 & w2v-last & 49.57 & 55.71 & 40.73 & 45.40 & 43.77 & 48.79 & -\\
 & w2v-avg & 48.17 & 54.36 & 39.81 & 44.20 & 42.86 & 47.84 & 47.79\\
\midrule
\multirow{4}{*}{\parbox{2.75cm}{tfcoco+tfoid+ places365+color}} & no filter & 78.64 & 78.98 & 5.45 & 22.51 & 30.34 & 21.96 & -\\
 & w2v-max & 50.24 & 57.88 & 38.19 & 45.75 & 43.09 & 48.69 & 48.91\\
 & w2v-last & 50.93 & 57.81 & 41.04 & 46.41 & 44.13 & 50.36 & -\\
 & w2v-avg & 50.47 & 58.10 & 39.33 & 45.78 & 42.93 & 49.61 & 49.51\\
\midrule
\multirow{4}{*}{\parbox{2.75cm}{tfcoco+tfoid+ places365+yolo9000+ color}} & no filter & 79.10 & 79.44 & 5.69 & 22.53 & 30.34 & 18.63 & -\\
 & w2v-max & 48.59 & 56.05 & 37.09 & 44.44 & 42.10 & 47.21 & 47.45\\
 & w2v-last & 49.17 & 55.81 & 39.97 & 45.00 & 43.08 & 48.74 & -\\
 & w2v-avg & 48.01 & 55.53 & 37.86 & 43.81 & 41.52 & 47.63 & 47.61\\
 \bottomrule    
    \end{tabular}
  \end{center}
  \caption{Non-paired phrase localization accuracies for Flickr30kEntities.}
  \label{tbl:fullaccuracy-flickr30k}
\end{table*}

\begin{table*}
  \begin{center}
  	\small
  	\begin{tabular}{l l |cc| ccccc}
    \toprule
    \multirow{2}{*}{\textbf{Detector}} & \multirow{2}{*}{\textbf{Similarity}} & \multicolumn{2}{c|}{\textbf{Upperbound}} & \multicolumn{5}{c}{\textbf{Localization strategy}} \\
        & & -union & +union & random & largest & confidence & union & consensus\\
\midrule
\multirow{4}{*}{\parbox{2.75cm}{tfcoco20}} & no filter & 27.21 & 30.33 & 8.22 & 13.20 & 11.93 & 12.91 & -\\
 & w2v-max & 24.35 & 26.82 & 11.01 & 14.97 & 14.31 & 13.97 & -\\
 & w2v-last & 19.48 & 21.82 & 9.83 & 13.26 & 12.39 & 13.26 & -\\
 & w2v-avg & 24.20 & 26.64 & 10.77 & 14.92 & 14.24 & 13.92 & -\\
\midrule
\multirow{4}{*}{\parbox{2.75cm}{tfcoco}} & no filter & 31.84 & 35.18 & 7.15 & 13.35 & 11.79 & 13.00 & -\\
 & w2v-max & 25.00 & 27.16 & 11.49 & 15.40 & 14.73 & 13.85 & -\\
 & w2v-last & 17.13 & 19.08 & 9.55 & 12.57 & 11.73 & 12.51 & -\\
 & w2v-avg & 24.43 & 26.54 & 11.25 & 15.24 & 14.55 & 13.73 & -\\
\midrule
\multirow{4}{*}{\parbox{2.75cm}{tfoid}} & no filter & 32.75 & 36.85 & 10.55 & 14.82 & 12.93 & 14.61 & -\\
 & w2v-max & 25.46 & 28.03 & 18.48 & 19.82 & 19.87 & 18.88 & -\\
 & w2v-last & 19.98 & 22.36 & 15.23 & 16.49 & 16.24 & 16.52 & -\\
 & w2v-avg & 25.13 & 27.73 & 18.35 & 19.68 & 19.71 & 18.89 & -\\
\midrule
\multirow{4}{*}{\parbox{2.75cm}{yolo9000}} & no filter & 9.75 & 13.22 & 6.71 & 8.37 & 7.09 & 10.34 & -\\
 & w2v-max & 7.93 & 9.56 & 7.32 & 7.64 & 7.45 & 8.95 & -\\
 & w2v-last & 7.76 & 9.30 & 7.24 & 7.50 & 7.34 & 8.78 & -\\
 & w2v-avg & 7.93 & 9.57 & 7.32 & 7.63 & 7.44 & 8.95 & -\\
\midrule
\multirow{4}{*}{\parbox{2.75cm}{places365}} & no filter & 14.64 & 14.64 & 14.64 & 14.64 & 14.64 & 14.64 & -\\
 & w2v-max & 14.64 & 14.64 & 14.64 & 14.64 & 14.64 & 14.64 & -\\
 & w2v-last & 14.64 & 14.64 & 14.64 & 14.64 & 14.64 & 14.64 & -\\
 & w2v-avg & 14.64 & 14.64 & 14.64 & 14.64 & 14.64 & 14.64 & -\\
\midrule
\multirow{4}{*}{\parbox{2.75cm}{color}} & no filter & 40.43 & 45.83 & 2.04 & 14.74 & 7.92 & 14.52 & -\\
 & w2v-max & 13.92 & 20.52 & 5.82 & 11.08 & 8.49 & 13.99 & -\\
 & w2v-last & 10.69 & 16.85 & 4.70 & 8.78 & 6.79 & 12.41 & -\\
 & w2v-avg & 13.82 & 20.58 & 5.93 & 11.05 & 8.61 & 14.10 & -\\
\midrule
\multirow{4}{*}{\parbox{2.75cm}{tfcoco+tfoid}} & no filter & 42.73 & 46.33 & 7.76 & 15.00 & 12.36 & 14.39 & -\\
 & w2v-max & 30.35 & 33.10 & 18.06 & 21.21 & 20.69 & 19.45 & 20.10\\
 & w2v-last & 20.95 & 23.39 & 13.77 & 16.53 & 15.49 & 16.44 & -\\
 & w2v-avg & 29.99 & 32.70 & 17.78 & 21.21 & 20.55 & 19.38 & 19.68\\
\midrule
\multirow{4}{*}{\parbox{2.75cm}{tfcoco+tfoid+ yolo9000}} & no filter & 43.00 & 46.63 & 7.13 & 14.01 & 12.24 & 13.15 & -\\
 & w2v-max & 28.44 & 31.13 & 17.04 & 19.88 & 19.54 & 18.40 & 19.30\\
 & w2v-last & 18.87 & 21.29 & 12.80 & 15.21 & 14.35 & 15.48 & -\\
 & w2v-avg & 26.99 & 29.70 & 16.56 & 19.46 & 19.15 & 18.14 & 18.53\\
\midrule
\multirow{4}{*}{\parbox{2.75cm}{tfcoco+tfoid+ places365}} & no filter & 48.80 & 49.14 & 8.76 & 14.65 & 12.53 & 14.58 & -\\
 & w2v-max & 33.00 & 35.45 & 20.85 & 23.93 & 23.57 & 21.97 & 22.64\\
 & w2v-last & 22.84 & 24.98 & 16.40 & 18.81 & 17.85 & 18.67 & -\\
 & w2v-avg & 32.61 & 35.04 & 20.93 & 23.95 & 23.43 & 21.95 & 22.25\\
\midrule
\multirow{4}{*}{\parbox{2.75cm}{tfcoco+tfoid+ places365+yolo9000}} & no filter & 49.08 & 49.42 & 8.03 & 14.53 & 12.50 & 13.52 & -\\
 & w2v-max & 31.07 & 33.55 & 19.77 & 22.55 & 22.34 & 20.94 & 21.83\\
 & w2v-last & 20.98 & 23.16 & 15.41 & 17.57 & 16.77 & 17.76 & -\\
 & w2v-avg & 29.53 & 32.03 & 19.50 & 22.08 & 21.87 & 20.64 & 21.01\\
\midrule
\multirow{4}{*}{\parbox{2.75cm}{tfcoco+tfoid+ places365+color}} & no filter & 69.78 & 70.00 & 4.21 & 14.81 & 12.58 & 14.49 & -\\
 & w2v-max & 35.57 & 39.50 & 21.22 & 26.48 & 25.36 & 24.52 & 25.52\\
 & w2v-last & 23.82 & 27.24 & 15.95 & 20.05 & 18.36 & 20.49 & -\\
 & w2v-avg & 34.91 & 38.97 & 20.96 & 26.45 & 24.92 & 24.66 & 25.11\\
\midrule
\multirow{4}{*}{\parbox{2.75cm}{tfcoco+tfoid+ places365+yolo9000+ color}} & no filter & 69.93 & 70.13 & 4.02 & 14.78 & 12.58 & 13.43 & -\\
 & w2v-max & 34.19 & 38.07 & 20.45 & 25.44 & 24.51 & 23.69 & 24.86\\
 & w2v-last & 22.76 & 26.13 & 15.57 & 19.23 & 17.76 & 19.82 & -\\
 & w2v-avg & 32.51 & 36.52 & 20.01 & 24.98 & 23.78 & 23.57 & 24.08\\
 \bottomrule
    \end{tabular}
  \end{center}
  \caption{Non-paired phrase localization accuracies for ReferItGame.}
  \label{tbl:fullaccuracy-refclef}
\end{table*}
\newpage

\end{document}